\pdfoutput=1
\PassOptionsToPackage{table,xcdraw}{xcolor} 

\documentclass[11pt]{article}

\usepackage[final]{acl}

\usepackage{times}
\usepackage{latexsym}

\usepackage[T1]{fontenc}

\usepackage[utf8]{inputenc}

\usepackage{microtype}

\usepackage{inconsolata}

\usepackage[most]{tcolorbox}
\usepackage[most]{tcolorbox}
\usepackage{siunitx}
\usepackage[utf8]{inputenc}
\usepackage{geometry}
\usepackage[table,xcdraw]{xcolor} 
\usepackage{array}                
\usepackage{booktabs}             
\usepackage{longtable}            
\usepackage{ragged2e}             
\usepackage{graphicx}             
\usepackage{amsmath, amsfonts, amssymb} 
\usepackage{enumitem}            
\usepackage{xurl}                 
\usepackage{tabularx}             
\usepackage{calc}                
\usepackage{multirow}             
\usepackage{xspace}               
\usepackage{makecell}             
\usepackage{pdfpages}             
\usepackage{hyperref}             
\usepackage{tabularray}
\usepackage{hhline}
\usepackage{rotating}
\usepackage{diagbox}
\usepackage{adjustbox}
\usepackage{subcaption} 
\usepackage{tablefootnote}
\usepackage{caption}
\usepackage{float}

\usepackage{enumitem}

\usepackage{siunitx}      
\usepackage{etoolbox} 
\usepackage{pifont}

\newcolumntype{Y}{>{\centering\arraybackslash}X} 
\newcolumntype{L}{>{\raggedright\arraybackslash}p{0.21\textwidth}} 
\newcolumntype{D}{>{\raggedright\arraybackslash}p{1.7cm}}          
\newcolumntype{T}{>{\raggedright\arraybackslash}X}                 
\newcolumntype{R}[1]{>{\raggedleft\arraybackslash}p{#1}}           

\newcommand{\name}{\textsc{FaStfact}\xspace}
\newcommand{\metric}{$F_{1}@K'$\xspace}

\title{\name: Faster, Stronger Long-Form Factuality Evaluations in LLMs}

\author{
  Yingjia Wan$^{1}$,
  Haochen Tan$^{2}$,
  Xiao Zhu$^{3}$, 
  Xinyu Zhou$^{3}$, 
  Zhiwei Li$^{3}$, 
  Qingsong Lv$^{5}$,\\
  {\bf Changxuan Sun$^{6}$,
  Jiaqi Zeng$^{7}$,
  Yi Xu$^{8}$,
  Jianqiao Lu$^{9}$,
  Yinhong Liu$^{10 *}$,
  Zhijiang Guo$^{3, 4}$}\thanks{Corresponding authors.} \\
  $^{1}$UCLA,
  $^{2}$CUHK, 
  $^{3}$HKUST (GZ), 
  $^{4}$HKUST,
  $^{5}$Tsinghua University, \\
  $^{6}$ECNU,
  $^{7}$NVIDIA,
  $^{8}$UCL, 
  $^{9}$HKU, 
  $^{10}$University of Cambridge \\
  \texttt{alisawan@ucla.edu} \\
  \texttt{yl535@cam.ac.uk}, 
  \texttt{zhijiangguo@hkust-gz.edu.cn} 
}

\begin{document}
\maketitle

\begin{abstract}
Evaluating the factuality of long-form generations from Large Language Models (LLMs) remains challenging due to efficiency bottlenecks and reliability concerns. Prior efforts attempt this by decomposing text into claims, searching for evidence, and verifying claims, but suffer from critical drawbacks: (1) inefficiency due to overcomplicated pipeline components, and (2) ineffectiveness stemming from inaccurate claim sets and insufficient evidence. To address these limitations, we 
propose \textbf{\name, an evaluation framework that achieves the highest alignment with human evaluation and time/token efficiency among existing baselines.} \name first employs chunk-level claim extraction integrated with confidence-based pre-verification, significantly reducing the time and token cost
while ensuring reliability. For searching and verification, it collects document-level evidence from crawled web-pages and selectively retrieves it during verification.
Extensive experiments based on an annotated benchmark \textbf{\name-Bench} demonstrate the reliability of \name in both efficiently and effectively evaluating long-form factuality. Code, benchmark data, and annotation interface tool are available at \url{https://github.com/Yingjia-Wan/FastFact}.
\end{abstract}
\begin{figure*}[t!]
    \centering
    \includegraphics[width=\linewidth]{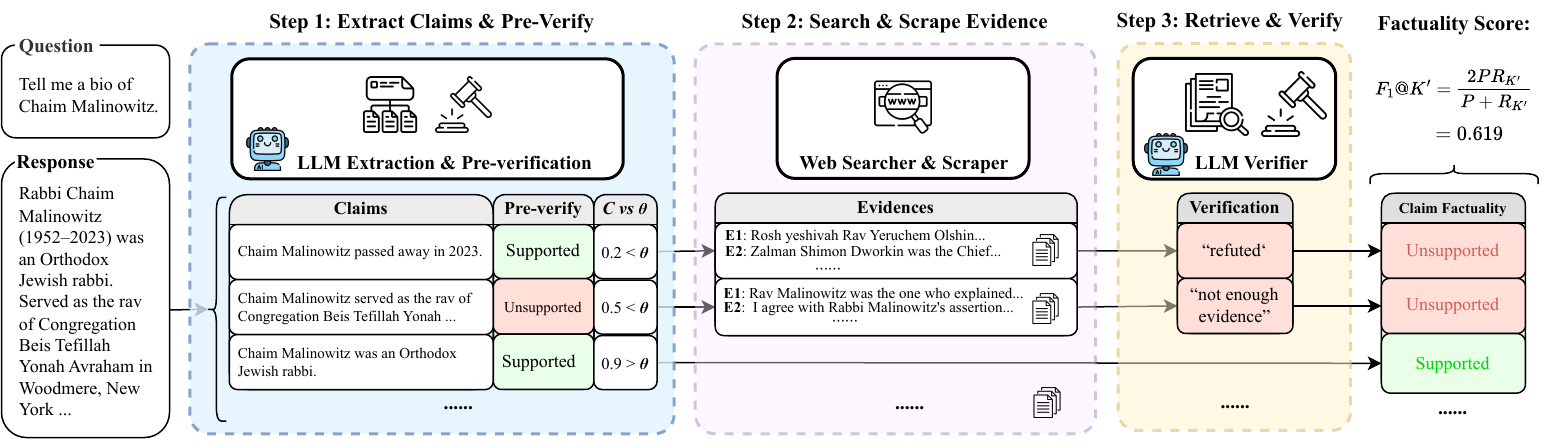}
    \caption{Overview of \name. \textbf{Step 1}: An extractor receives a chunk of a long-form generation to extract and pre-verify atomic claims. For each claim, if the pre-verification confidence $C$ falls below a predefined threshold $\theta$, proceed to step 2 and 3; otherwise, skip the following steps and use pre-verification for claim factuality. \textbf{Step 2}: A web search is triggered to crawl document-level evidence for each claim. \textbf{Step 3}: Based on the gathered evidence, a verifier classifies the factual correcteness of each claim. A final factuality score \metric (\autoref{subsec: metric}) is then computed by aggregating the claim-level verifications.}
    \label{fig:pipeline}
\end{figure*}
\begin{table*}[t]
\centering
\small
\captionsetup{width=\linewidth}

\begin{tabularx}{\textwidth}{l*{5}{>{\arraybackslash}X}}
\toprule
\textbf{Evaluation Pipelines} &
  \textbf{\name} &
  {ExpertQA} &
  FacTool &
  {VeriScore} &
  {SAFE} \\
\midrule
$\text{F1@K}'$ (Ground-truth $\text{F1@K}'$= 0.792) & 0.780 & 0.919 & 0.987 & 0.899 & 0.960 \\
$|\Delta\text{F1@K}'|$ (Score Alignment) & \textbf{0.012} & 0.127 & 0.195 & 0.107 & 0.168 \\
$|\Delta K|$ (Number of Claims Alignment) & \textbf{3.35} & 21.26 & 9.09 & 7.32 & 14.28 \\
Avg Token Cost & 5615 & 7893 & 4480 & 22848 & 49622 \\
\bottomrule
\end{tabularx}
\caption{Baseline comparison on \name-Bench in terms of ground-truth alignment and token cost. ExpertQA is directly imported from FactSCore, except augmented with Google Search.
We implement FacTool in the knowledge-based QA setting to be compatible with the domains of \name-Bench.
}
\label{tab:baseline-compare}
\vspace{-3mm}
\end{table*}
\vspace{-3mm}
\section{Introduction}
\label{sec:intro}

Despite continuous advancements of Large Language Models (LLMs) in downstream tasks ~\citep{DeepSeekV3,Qwen2,o1,Gemini2.0Pro}, a significant challenge remains in their ability to reliably answer in-depth factuality questions. A particular concern is the tendency of LLMs to generate factual errors where claims are contradicted or unverified by established ground-truth knowledge, a problem especially prevalent in long-form generations~\citep{ExpertQA2024, factcheck-bench}. 

Existing methods for evaluating long-form generations, such as FActScore~\citep{factscore} and SAFE~\citep{SAFE_wei2024}, typically follow a three-stage pipeline involves: (1) decomposing the generation into a list of atomic claims; (2) retrieving relevant evidence for each claim, often from sources like Wikipedia or Google Search; and (3) verifying each claim against the retrieved evidence to determine its factuality.

In this paper, we first analyze prior factuality evaluation methods and find that existing pipelines suffer from significant drawbacks in both efficiency and effectiveness, substantially hindering their applicability (\autoref{sec:analysis}). From a systematic analysis, inefficiency stems from the sentence-level decomposition and complex decontextualization during claim extraction, which subsequently harms the later verification steps. The high processing time and inference costs significantly limit their practical use, particularly for evaluating the increasingly lengthy outputs of modern LLMs. In terms of ineffectiveness, existing evaluations often yield inaccurate or incomplete claim sets. They mistakenly include unverifiable or redundant claims while missing essential information. These fundamental failures primarily result from the inherent pipeline design, undermining subsequent verification and degrading overall evaluation quality. Furthermore, existing methods often struggle with insufficient evidence retrieval, frequently relying on short, inadequate snippets that prevent conclusive verification even when ample online information exists.

To address the limitations of both inefficiency and ineffectiveness, we then propose \textbf{\name} (\autoref{sec:method}), a more efficient and effective framework for evaluating the factuality of long-form responses. \name begins with claim extraction and pre-verification, where an LLM extracts atomic claims from the response and pre-verify them. To combat the inefficiency of prior methods, this step utilizes dynamic chunking for longer context and accelerated processing, and incorporates a confidence-based pre-verification leveraging the LLM's internal knowledge, thus reducing the need for external verification on straightforward claims. To sufficiently fact-check the claims that are pre-verified with high uncertainty, \name proceeds to collect document-level evidence via web-scraping. This design directly addresses the issue of insufficient evidence by fetching entire web page contents from search results to create a comprehensive knowledge base.

To reliably assess \name in comparison with existing baselines, we further aggregate and annotate a benchmark \textbf{\name-Bench} (\autoref{sec:benchmarks}) consisting of 400 pairs of factually annotated long-form QA. The benchmark provides sub-process ground-truths such as number of extracted claims, and specific verification label per claim, which enables us to assess \name's alignment with human judgment, followed by a systematic baseline comparison with existing evaluation tools. \name shows superiority in processing time (\autoref{sec: efficiency}), token cost, and gap from ground-truth scores (\autoref{tab:baseline-compare}), rendering it a more efficient, effective, and reliable tool for the factuality evaluation of long-form generation.

\section{Related Work}
\label{sec:related}

LLMs exhibit factual errors, particularly in open-ended QA responses, leading to the development of factuality evaluation benchmarks. While early efforts utilized short-form benchmarks like SimpleQA \citep{SimpleQA}, TruthfulQA \citep{TruthfulQA}, and HalluQA \citep{HalluQA2023} to assess individual factoids based on answer-matching correctness, evaluating the factuality of complex long-form generations presents greater challenges. The prevailing approach for long-form factuality evaluation is the \textbf{decompose-then-verify} framework \citep{factscore, SAFE_wei2024,veriscore, factool_chern2023, factcheck-bench, ExpertQA2024, OpenFactCheck}. 
Systems like FActScore \citep{factscore} pioneered this framework for biographies using a closed-domain corpus, while SAFE \citep{SAFE_wei2024}, ExpertQA \citep{ExpertQA2024}, and Factcheck-Bench \citep{factcheck-bench} extended evidence gathering to open-domain search. VeriScore \citep{veriscore} refined claim extraction and context engineering, and FacTool \citep{factool_chern2023} broadened evaluation domains to include reasoning, coding, and scientific literature by incorporating diverse evidence sources like Python executors and Google Scholar. Additional related work on factuality benchmarks is in~\autoref{appdx:additional_related}.
\section{Analysis of Existing Long-Form Factuality Evaluation Systems}
\label{sec:analysis}

This section analyzes current factuality evaluation pipelines for long-form content within the decompose-then-verify framework that follows a static agent workflow: (1) \textit{Claim Extraction}: decompose long-form content into atomic claims; (2) \textit{Evidence Retrieval}: retrieve evidence pertinent to each claim; (3) \textit{Claim Verification}: verify the factuality of each claim based on the gathered evidence; and (4) \textit{Overall Scoring}: combine factuality of each claim into an overall factuality score. We examine their limitations in terms of efficiency, effectiveness, and reliability compared to human judgment.



\subsection{Inefficient Claim Extraction}
Existing factuality evaluation pipelines are inefficient in token cost and processing time, hurting their applicability and scalability to document-level longer generations. 
Such inefficiency can be mainly attributed to two specific designs within the claim extraction step:

\paragraph{Checks and Revisions.}
Existing evaluation pipelines~\citep{factscore,SAFE_wei2024, ExpertQA2024, factcheck-bench} perform checks and revisions after decomposing each sentence in the response, which contributes to significant inefficiency. For instance, SAFE \citep{SAFE_wei2024} subjects each factual claim to both a revision and a relevance check to ensure the claim is context-independent and pertinent to the question topic. Both steps necessitate sequential inference calls. Furthermore, SAFE's reliance on chain-of-thought prompting for all LLM-based operations substantially increases both token cost and inference latency. Similarly, \citet{factcheck-bench} includes decontextualization and checkworthiness assessment as post-processing steps following the initial claim decomposition. As the capabilities of mainstream extractor models have advanced tremendously from their predecessors (e.g., InstructGPT in FactSCore), these claim checks and revisions have become a redundant legacy practice that adds inference costs and processing time.


\paragraph{Sentence-Level Claim Extraction.}
The other major contributor to significant computational overhead is the current default of sentence-level claim extraction, which necessitates massive inference calls. A long-form generation is parsed into single sentences for the claim extractor to extract factual claims from each sentence independently. While this approach offers comprehensive coverage by attending to all parts of responses, its computational cost—in time and tokens—scales linearly with the length of the generated text.

Beyond inefficiency, limiting the extraction scope to each single sentence also causes isolation and insufficient context, causing failures discussed in \autoref{subsec:failures in claim extraction}. While VeriScore attempts to mitigate this with sliding windows, its extractor still operates per-sentence, retaining the same time complexity and largely restricted context. 


\subsection{Ineffective Claim Extraction}
\label{subsec:failures in claim extraction}


Issues in claim extraction, a critical first step in the hierarchical decompose-then-verify framework, can significantly compromise subsequent evaluation steps and overall assessment quality. Through case studies and statistical analyses, we show that the aforementioned practices in existing evaluation pipelines not only add considerable inference cost and processing time, but also introduce systematic failures, which negatively impact the quality of claim extraction and the overall evaluation. Fully detailed analyses are attached in ~\autoref{appdx:parent-claim-extraction-failure-analyses}.


We first take a case inspection of SAFE by manually assessing its extracted claims from a GPT-4 generation on a random question from LongFact~\citep{SAFE_wei2024}. This reveals a total of 39 problematic claim extraction cases among all 57 claims decomposed from a GPT-3.5-Turbo generation of 19 sentences long, rendering a striking 68\% failure rate. To aid a systematic analysis, we identify three major categories of claim extraction failures in existing factuality evaluations and their subcategories below. Appendix~\ref{appdx:failure-cases-SAFE} provides a case study illustrating examples of each failure category sampled from SAFE.

\noindent \textbf{(A) Unverifiable Claims} are statements that fail to describe a single event or state with necessary modifiers (e.g., spatial, temporal, or relative clauses). Existing pipelines like FActScore and SAFE are found to produce unverifiable claims in three types: 

\textit{1) Subjective}: claims about a story, personal experiences, hypotheticals (e.g., ``would be'' or subjunctive), subjective statements (e.g., opinions), future predictions, suggestions, advice, instructions, etc. (e.g., claim 3 in Appendix~\ref{appdx:failure-cases-SAFE});

\textit{2) Tautology}: claims that restate themselves and carry no meaning (e.g., claim 12 in Appendix~\ref{appdx:failure-cases-SAFE});

\textit{3) Ambiguous}: claims that contain equivocal information or that are contextually dependent (e.g., not situated within relevant temporal information and location, or no clear referent for entities) (e.g., claim 1,2 in Appendix~\ref{appdx:failure-cases-SAFE}).

\noindent \textbf{(B) Redundant Claims} are repeated statements that cover identical information. Current metrics are vulnerable to extracting redundancy below:

\textit{1) Intra-sentence redundancy}: semantically overlapping claims extracted from the same sentence during a single inference pass. This occurs due to (a) the restrictive context window (sentence-level extraction) and (b) inadequate deduplication processes in checks and revisions. For example, claims 4 \& 5 in Appendix~\ref{appdx:failure-cases-SAFE} describe near-identical statements except minor phrasing nuances.
    
\textit{2) Inter-sentence redundancy}: semantically overlapping claims extracted across different sentences in the generation. This is a system weakness in existing pipelines that fundamentally stems from their sentence-level context window, which prevents the extractor from recognizing global generation structure (e.g., claims 8 \& 9 restate claims 4 \& 5 in Appendix~\ref{appdx:failure-cases-SAFE}). It occurs prevalently in long-form generation evaluation, such as articles with an overview-body-summary structure, with an escalating frequency as the length increases.

\noindent \textbf{(C) Missing Claims} are verifiable statements that should have been extracted, but current efforts failed due to two circumstances:

\textit{1) Removed by Relevance Checks}: Post-hoc relevance checks and revisions unexpectedly label valid claims as irrelevant and remove them, (e.g., claim 13 in Appendix~\ref{appdx:failure-cases-SAFE}).

\textit{2) Missed by the Extractor}: The extractor occasionally fail to extract a part of valid claims. This can be attributed to two causes: (a) system weakness of the framework design: sentence-level claim decomposition makes it impossible to identify inter-sentence claims (e.g., claim 14 in Appendix~\ref{appdx:failure-cases-SAFE}); (b) model weakness in instruction following as the extractor, affected by prompt design.



Based on the categorical analysis on the case study, another preliminary analysis is conducted on a more recent pipeline VeriScore \citep{veriscore} to investigate the distribution of each failure category. We utilize an LLM-as-judge to classify the failure types of extracted claims, adapting from \citet{Decomposition_Dilemmas}'s analysis method. The detailed analyses of VeriScore distributions of claim extraction failures are provided in Appendix~\ref{appdx:failure-distribution-analysis-of-veriscore}.

Through case inspection and distribution analyses, we find that a major proportion of claim extraction failures stems from structural weaknesses in widely used evaluation pipelines. For instance, restricting the extractor’s context window to a single sentence prevents it from capturing inter-sentence claims that require integrating information across multiple sentences or paragraphs in long-form text. Moreover, the checks and revisions introduced during extraction often induce over-decomposition, producing redundant or unverifiable outputs such as tautologies and ambiguous claims..

\subsection{Unreliable Evidence-Based Verification}
\label{subsec:failures in evidence}
Each extracted claim needs to be verified for its factual correctness against external world knowledge by retrieving evidence from either a closed-domain knowledge source or Google Search. The retrieved evidence will be appended as context for the verifier to classify the final factuality label per claim.

Current search-augmented evaluations empowered by Google Search provide only limited evidence for the verifier \citep{SAFE_wei2024, factcheck-bench, veriscore, factbench-Bayat2024}. A common practice is to use the short snippet (a short description about the link content in 20–40 tokens) returned with each search result via the Serper API\footnote{https://serper.dev/} as the sole evidence, rather than the richer content on the linked webpages. These truncated, incomplete sentences offer little meaningful context for fact-checking, often forcing verifiers into inconclusive decisions even when sufficient information exists online. \autoref{appdx:cases-evidence-search} illustrates such a failure case from VeriScore \citep{veriscore}.

\section{\name}
\label{sec:method}

\begin{table*}[ht]
\vspace{-0.6em}
\centering
\small

\begingroup 
\setlength{\tabcolsep}{4pt} 
\resizebox{\linewidth}{!}{
\begin{tabular}{@{}llrrrrrr@{}}
\toprule
\textbf{Task/Benchmark} & \textbf{Statistics} & \textbf{Overall} & \textbf{ExpertQA} & \textbf{FactcheckBench} & \textbf{FActScore\_bio} & \textbf{HelloBench} & \textbf{LongFact} \\
\midrule
Claim Extraction & \# removed & 1.233 & 1.173 & 0.872 & 0.605 & 2.100 & 1.382 \\
 & \# added & 2.171 & 1.346 & 1.064 & 1.145 & 3.900 & 3.395 \\
 & \# revised & 0.928 & 1.679 & 0.167 & 0.605 & 0.875 & 1.289 \\
 & \# unchanged & 14.829 & 14.284 & 9.769 & 8.658 & 19.150 & 22.224 \\
\midrule
Claim Verification & agreement rate (exact) & 85.3\% & 94.7\% & 77.8\% & 68.7\% & 95.3\% & 88.8\% \\
  & agreement rate (type) & 92.9\% & 95.6\% & 91.8\% & 87.5\% & 96.0\% & 93.4\% \\
\midrule
Unit Alignment & $|\Delta  K'|$ & 3.350 & 2.506 & 1.949 & 1.882 & 5.688 & 4.697 \\
 & $|\Delta  F_1@K'|$ & 0.012 & 0.043 & 0.026 & 0.156 & 0.114 & 0.025 \\
 & $|\Delta  \text{\#supported} |$ & 3.675 & 2.457 & 2.282 & 2.645 & 5.475 & 5.539 \\
 & $|\Delta  \text{\#unsupported} |$ & 1.412 & 0.543 & 1.295 & 2.329 & 0.825 & 2.158 \\
 & $|\Delta  \text{\#irrelevant} |$ & 0.041 & 0.025 & 0.051 & 0.013 & 0.062 & 0.053 \\
\midrule
End-to-End Alignment & \metric $  $ (Ground-truth) & 0.792 & 0.901 & 0.802 & 0.522 & 0.889 & 0.833 \\
 & \metric $  $ (\name) & 0.780 & 0.858 & 0.776 & 0.677 & 0.774 & 0.808 \\
\bottomrule
\end{tabular}
}
\endgroup
\caption{Human-\name alignment in each stage of the pipeline and end-to-end scores. For claim verification (annotation task 2), the agreement rate between human and \name verification is calculated in two ways: \textit{exact} label matching among [``supported'', ``not enough confidence'', ``conflicting evidence'', ``refuted'', ``irrelevant'']; \textit{type} matching in terms of [``supported'', ``non-supported'', ``irrelevant''], which affects the final scoring.}
\vspace{-0.6em}
\label{tab:annotation_delta}
\end{table*}

To address the limitations of existing pipelines, we propose \name, a factuality evaluation framework of novel pipeline and metric design for efficient assessment of long-form generation. 

\subsection{Claim Extraction \& Pre-Verification}


\paragraph{Chunk Window with Configurable Stride:}
To ease the significant time and token overhead during the claim extraction step, when preparing the long-form QA as input for the LLM extractor, we replace the sentence-level sliding window with a flexibly configurable chunk window. The chunk stride $w$ (i.e., number of sentences within a chunk) is configurable between a minimum of 1 (working equivalently as a sentence-level sliding window), and a maximum of the entire response length. This brings several benefits: firstly, it scales the efficiency of the claim extraction process by reducing the number of inference calls for decomposition; secondly, a longer context window empowers the extractor to be more immune to extracting (both inter-sentence and intra-sentence) redundant claims and unverifiable claims, as supported by later experiments. Moreover, the configurable hyperparameter of chunk stride enables us to tune the granularity of the entire factuality evaluation. This helps to customize a user's need when employing the \name tool, to achieve a balance in the tradeoff between claim quality, efficiency, and ultimately the accuracy of the factuality score in effectively aligning with human judgment.


\paragraph{Verifiable Claim Extraction:}
Instead of post-hoc claim checks and revisions, we task the LLM extractor to handle these together in extraction in one inference, by following \citet{veriscore}'s guideline of extracting verifiable atomic claims. Through explicit instructions and few-shot demonstrations, the extractor can generate verifiable atomic claims based on the chunked response and its corresponding question for reference. The prompt template is provided in~\autoref{appdx:prompt-extraction}.

\paragraph{Confidence-Based Pre-Verification:} 
To further address the inefficiency limitations, we exploit LLMs' internal knowledge to verify simple claims where external evidence is not necessary. Based on the decompose-then-verify framework, a valid atomic claim can be categorized as a description about a specific event, an attribute about a specific entity, or a constant truth or principle \citep{factscore, veriscore}. An intuitive observation is that, among a set of claims, the verification difficulty varies from one another, with some requiring more concrete evidence and in-depth exploration than others. Similarly, for LLMs as a verifier, some widely known facts are easier to verify if there are abundant similar statements or relevant information in their training data, as empirically supported in investigations of reference-free fact-checking \citep{language-models-know-what-they-know, factscore}.

Leveraging this observation, \name\ adds \textbf{pre-verificatoin}, where it tasks the extractor model to classify its confidence in determining the relevance and factual correctness of each claim right after extraction during one-time inference, by choosing a label from [``Irrelevant'', ``Supported'', ``Non-supported'', ``Likely Supported'', ``Likely Non-supported'', ``Unsure''] via reference-free classification. We deliberately expanded the labels from the verification labels [``Irrelevant'', ``Supported'', ``Non-supported''] from existing works \citep{factscore, SAFE_wei2024}, to let the extractor verbally showcase uncertainty or indecisiveness, hence naturally filtering out claims necessitating further evidence-based verification. Otherwise, we yield the freedom to the extractor to act as a judge in verifying simple factual claims against its parametric knowledge, which spares subsequent steps and significantly boosts efficiency.

To curb overconfidence in pre-verification (e.g., the extractor assertively labeling all claims as true without discrimination), we incorporate \textbf{confidence checking} as a monitoring mechanism, inspired by studies in predictive uncertainty \citep{predictive-uncertainty-controled-generation} and generative verifiers using logprobs in next-token prediction \citep{generative_verifiers/ZhangHBKKA25}. A pre-verification label is accepted as valid for skipping the subsequent evidence searching and verifying steps only if both the following conditions are met:

1. The pre-verification label is definite (i.e., ``Supported'', ``Non-supported'', or ``Irrelevant'').

2. The label token's predictive confidence \( C \), defined as the normalized log probabilities of the label token, is higher than a calibrated threshold value\footnote{\name's robustness to confidence threshold calibration is discussed in \ref{sec:ablations} Ablations.}. Formally, given a claim \( x \) and its label \( y \in \{\text{Supported}, \text{Non-supported}, \text{Irrelevant}\} \): 

\vspace{-10pt}
\begin{align}
    C = \exp(\text{target\_logprob}) = P_\theta(y \mid x),
\end{align} where: \(\text{target\_logprob} = \log P_\theta(y \mid x)\) is the log probability of the label token \( y \), as directly returned by the extractor model. \( P_\theta(y \mid x) \) is softmax over model logits. \(\theta\) denotes the model parameters.

A higher value of \( C \) indicates greater confidence in the model's pre-verification decision. Configuring confidence checking allows greater control and flexibility in factuality evaluation, enabling a balance between efficiency and custom budget in response and search tokens.

\subsection{Document-Level Evidence Search}

For extracted claims that fail the pre-verification confidence check, we use them as queries to search the open web for relevant evidence to support or refute the claim. Instead of collecting short SERPER snippets regarding the query (consisting of a title, URL link, and a description snippet) in previous pipelines, we access each searched URL link and fetch the entire web page content\footnote{We use the Jina Reader API: https://jina.ai/reader/}. Therefore, each scraped content is a complete document listing coherent information regarding the webpage title. The average length of the scraped webpage document is 7054.29 words per piece, in contrast with the previous short snippet of 23.75 words that is frequently cut off as incomplete. This provides a much longer and grounded context serving as an informative knowledge base, which can later assist claim verification significantly. Specifically, as shown in the case comparison in Appendix~\ref{subsec:failures in evidence}, it addresses two limitations in previous evaluation pipelines: (1) proportion of `inconclusive' or `not enough evidence' labels from the verifier due to insufficient reference; (2) the overall verification reliability of the verifier model affected by the poor quality of searched snippets as evidence reference. The number of searched results is a configurable hyperparameter. Scraped web content is aggregated into a document-level knowledge base simulating a world knowledge source like Wikipedia corpora in FActScore~\citep{factscore}, but expanding to time-sensitive and dynamic open-domains.

\subsection{Retrieval-Augmented Verification}

During the claim verification stage, the verifier is provided with the searched evidence as a reference context in judging whether the target claim can be supported or refuted by the evidence. To avoid over-long input context and make the most efficient and effective use of the document-level knowledge base, we retrieve multiple chunks that are most relevant to the original claim using a BM2.5 retriever, and append them as evidence for the verifier. A context window or chunk is implemented to allow sentence integrity and more relevant information for the verifier to process. Hyperparameters include the overlapping length of each chunk, chunk length, number of searched webpage sources, and number of evidence pieces provided for the verifier. \autoref{appdx:cases-evidence-search} displays a success case of our evidence-based verification, juxtaposed with VeriScore's deficiency on the same claim.

For evidence-based verification labels, we adapt terminologies from established fact-checking works \cite{AVeriTeC, FIRE_factchecking}, asking the verifier to choose from [``supported'', ``refuted'', ``conflicting evidence'', ``not enough evidence'', ``unverifiable'']. 
The multi-class labeling setting bears two advantages: (1) \textit{transparency and robustness}: VeriScore uses ``inconclusive'' to encompass both cases of ``conflicting evidence'' and ``not enough evidence'', making it difficult to track verifier’s rationale; (2) \textit{backtracking}: ``unverifiable'' label offers a backtracking loop in assessing the previous extraction step. If an ambiguous or subjective claim accidentally gets passed through by the extractor, the verifier will flag it and drop it from the final claim list during score calculation. The two distinguished labels (``conflicting evidence'', ``not enough evidence'') are counted towards the ``non-supported'' for the final \name score calculation, hence compatible with existing metric settings. The prompt template for evidence-based claim verification is in~\autoref{appdx:prompt-verification}.

\subsection{\name\ Score Metric}
\label{subsec: metric}

\subsubsection{Background}

Early efforts, such as FActScore \citep{factscore}, utilize a simple \textbf{precision} metric that measures the factual accuracy of a response (i.e., the percentage of factually correct claims among all extracted claims). Given a response \(y\), let \(S(y)\) and \(N(y)\) respectively be the number of supported and non-supported facts. The factuality score for a long-form generation is:
\vspace{-1pt}
\begin{align}
P(y) = \frac{S(y) }{S(y) + N(y)} \label{eq:p}
\end{align}
\vspace{-5pt}

SAFE~\citep{SAFE_wei2024} extends the metric scope to cover both factual precision \(P(y)\) and recall \(R_{K}(y)\) (i.e., how many factually correct claims are covered in the response compared to those that \textit{should} be covered), formally denoted as:

\vspace{-10pt}
\begin{align}
F_1@K &= \frac{2P(y)R_{K}(y)}{P(y)+R_{K}(y)},
\label{eq:F_1}
\end{align}
\vspace{-10pt}

Here, a predefined $K$ is manually set to simulate the ``ideal'' number of supported facts from a QA response that counts as useful information for the question raiser, thus achieving full factual recall.
Hence:

\vspace{-20pt}
\begin{align}
R_{K}(y) &= \min\left(\frac{S(y)}{K}, 1\right) \label{eq:r_safe}
\end{align}
\vspace{-10pt}

SAFE's $F_1@K$ highlights an important aspect of factuality evaluation on the need to consider how the number of extracted claims can affect the effectiveness of the final score, making it a widely popular score metric \citep{veriscore, deepmind2024facts}. However, below we analyze the weaknesses of SAFE's $F_1@K$ by defying two of its disputable assumptions about long-form factuality evaluation:

\paragraph{Verbosity Blindspot:} Based on the definition of \(R_{K}(y)\), SAFE's $F_1@K$ metric penalizes only uninformative generations (i.e., \(\tfrac{S(y)}{K} < 1\)), but disregards overly long or verbose generations. In an extreme case where \(S(y) \gg K\), a model may continue producing factually trivial yet unhelpful content without its score being degraded. This limitation is particularly concerning given the prevalence of redundant claims in SAFE. While \citet{SAFE_wei2024} defends their design by arguing that LLMs do not repeat facts in their generations, this assumption has been empirically challenged by \citet{veriscore} and further contradicted by our analyses in \autoref{subsec:failures in claim extraction}, which reveal that redundant claims constitute the largest share of factual claim failures.
\paragraph{Arbitrary $K$ Value:} The critical parameter K suffers from circular dependency, as it is derived from auto-generated claims produced by the evaluation pipeline itself rather than ground truth annotations. Common practices of setting K as the median/maximum number of claims auto-extracted from sampled generations by the auto-evaluation pipeline make the metric contingent upon the evaluation systems and input generations. This methodological compromise introduces systematic bias, where benchmark performance becomes sensitive to both the extractor model bias and the quality distribution of the evaluated models rather than reflecting objective factuality standards.

\subsubsection{Metric Score Design}
Based on the analysis above, we propose the following \metric for \name:

\paragraph{Ground-truth \textbf{$K'$}} Rather than setting a artificially fixed $K$ averaged from sampled generations for the score metric, \name-Bench (\autoref{sec:benchmarks}) provides annotations of individual response-level $K'$s (i.e.,  the number of total claims that \textit{should} be extracted from each specific generation), enabling a more accurate estimation of factual recall.
\vspace{-3pt}
\paragraph{Symmetrical Penalty on Digressions from Ground-truth \textbf{$K'$}} We revise the factual recall formula from SAFE to impose symmetric penalties for both insufficient and excessive claim coverage: 

\vspace{-10pt}
\begin{equation}
R_{K'}(y) = \frac{2}{1 + e^{\gamma \left| S(y) - K' \right|}}
\label{eq:fastfact R}
\end{equation}

\vspace{-10pt}

\begin{align}
F_1@K' &= \frac{2P(y)R_{K'}(y)}{P(y)+R_{K'}(y)},
\label{eq:fastfact F_1}
\end{align}

\vspace{-5pt}

where \(\gamma\) is a scaling hyperparameter chosen to maintain a similar magnitude as \citet{SAFE_wei2024}. \autoref{eq:fastfact R} creates equal penalties on the number of verifiable claims \(K(y)\) extracted from a specific long-form generation if either exceeding \textit{or} falling short of the ground-truth \(K'(y)\). This design avoids the verbosity blindspot, which we emphasize based on the analyzed high distribution of claim redundancy in long-form factuality evaluation.

\begin{table*}

\setlength{\tabcolsep}{3pt}
\tiny
\centering

\vspace{-0.2em}


\resizebox{\linewidth}{!}{
\begin{tabularx}{\textwidth}
  {@{} L D T ccccc R{0.8cm} c >{\centering\arraybackslash}p{1.0cm} R{1.0cm} R{1.0cm}c R{0.6cm}c @{}}
\toprule
\textbf{Prompt Source} & \textbf{Domain} & \textbf{Task Type} &
\textbf{Prompt(Words)} & \textbf{\makecell[c]{Prompt(Sents)}} &
\textbf{\makecell[c]{Resp(Words)}} & \textbf{\makecell[c]{Resp(Sents)}} &
\textbf{Samples} \\[-0.20em]
\midrule
\makecell[l]{FActScore-Bio \citep{factscore}} &
Biography &
``Tell me the bio of ..." &
  7.1 & 1.05 & 222.8 & 11.2 & 80 \\

\makecell[l]{Factcheck-Bench \citep{factcheck-bench}} &
General &
Fact-seeking general QA &
 13.0 & 1.20 & 281.5 & 17.9 & 80 \\

\makecell[l]{ExpertQA \citep{ExpertQA2024}} &
Domain Expertise &
Domain-specific situational QA &
101.6 & 7.55 & 512.6 & 37.0 & 80 \\

\makecell[l]{LongFact \citep{SAFE_wei2024}} &
General &
Concept explanation; object description &
 47.0 & 2.45 & 586.6 & 36.0 & 80 \\

\makecell[l]{HelloBench \citep{hellobench}} &
General &
Open-ended QA; advice; essay writing &
115.1 & 11.00 & 721.8 & 51.6 & 80 \\

\midrule
\textbf{\name-Bench} & \textbf{(Aggregated)} & \textbf{(Aggregated)} &
\textbf{56.8} & \textbf{4.65} &
\textbf{465.1} & \textbf{30.8} &
\textbf{400} \\
\bottomrule
\end{tabularx}
}
\caption{Break-down of \name{} benchmark structure and statistics. Response lengths (words/sentences) are averaged over
DeepSeek-R1, DeepSeek-V3, GPT-4o and Qwen-2.5-Instruct.}
\label{tab:fastfact_bench_stats}
\end{table*}

\section{Efficiency Analysis}
\label{sec: efficiency}

To systematically compare the efficiency of \name with existing evaluation baselines, we offer a complexity analysis by summarizing the workflow of each pipeline.

\subsection{Key Parameters}
\begin{itemize}[noitemsep, parsep=0pt, labelwidth=1em, leftmargin=!]
    \item[$\mathbf{N}$:] Number of sentences in the input generation.
    \item[$\mathbf{M}$:] Number of atomic claims extracted from the $N$ sentences.
    \item[$\mathbf{q}$:] Custom number of search queries for evidence retrieval for each claim ($\mathbf{q} \ge 1$).
    \item[$\mathbf{p}$:] Ratio of extracted claims in \name whose pre-verification confidence $C$ did not pass the threshold $\theta$, triggering subsequent evidence retrieval and verification ($0 \le \mathbf{p} \le 1$).
    \item[${\mathbf{w}}$:] Custom chunk stride in \name i.e., the number of sentences provided to the LLM extractor for claim extraction ($1 \le \mathbf{w} \le N$).
\end{itemize}

\subsection{Inference Cost and Time Complexity}
\begin{figure}[htbp]
  \centering
    \includegraphics[width=\linewidth]{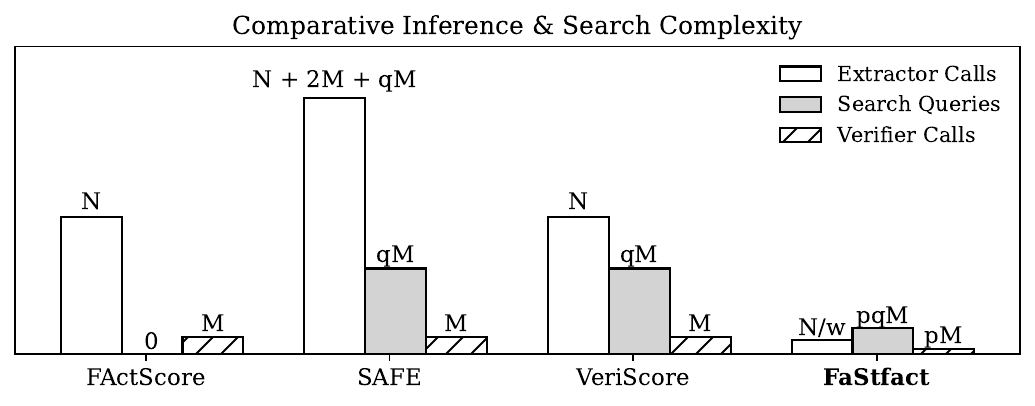}
    \vspace{-5pt}
    \caption{Breakdown of LLM Inference Calls and Search Queries by Pipeline Phase for different metrics.}
    \label{fig:complexity}
    \vspace{-5pt}
\end{figure}

The total cost and processing time of a factuality evaluation pipeline is dominated by two operations: LLM inference calls $O_{LLM}$ and online evidence search $O_{search}$, which both introduce latency and potential token cost. \autoref{fig:complexity} breaks down the operation complexity for each pipeline phase. \autoref{appdx:further-complexity-analysis} provides a detailed walkthrough regarding each table item for analyzing the efficiency of all baselines.


In our work, \name first reduces complexity by compressing extraction from $O_{LLM}(\mathbf{N})$ to $O_{LLM}(\mathbf{N}/\mathbf{w})$ via chunk-level processing. Meanwhile, by introducing confidence-based pre-verification, it prunes the claim set for further verification from $\mathbf{M}$ to $\mathbf{pM}$. As a result, both search and verification costs scale down to only $O_{search}(\mathbf{pqM})$ and $O_{LLM}(\mathbf{pM})$. This reduction makes \name substantially more efficient in both latency and token cost compared to prior pipelines, scaling sublinearly.

\section{\name-Bench}
\label{sec:benchmarks}

Following discussions in \autoref{sec:analysis}, despite numerous benchmarks targeting the evaluation of LLM long-form factuality, existing evaluations are scattered in their reported score metrics, tested benchmarks, and task domains, lacking a holistic baseline comparison. Many existing factuality benchmarks also lack ground-truth data for the sub-processes \citep{SAFE_wei2024, veriscore}, making it difficult to compare the reliability of factuality evaluation scores among pipelines in parallel. 

Therefore, we introduce \name-Bench aggregated over five representative benchmarks on LLM factuality in long-form generation and manually annotated them on both claim extraction and verification, enabling a comprehensive and robust baseline comparison on existing evaluation tools. \autoref{tab:annotation_delta} shows the fine-grained compositional human-alignment statistics of \name on \name-Bench, showing high accuracy both within claim extraction and verification and in the end-to-end scores; \autoref{tab:fastfact_bench_stats} lists the benchmark information across domains, tasks, and generation lengths. The benchmark construction and human annotation details are provided in~\autoref{sec:benchmark_construction}.

\section{Experiments}

\begin{figure*}[t]
\vspace{-7mm}
  \centering
  \includegraphics[width=\textwidth]{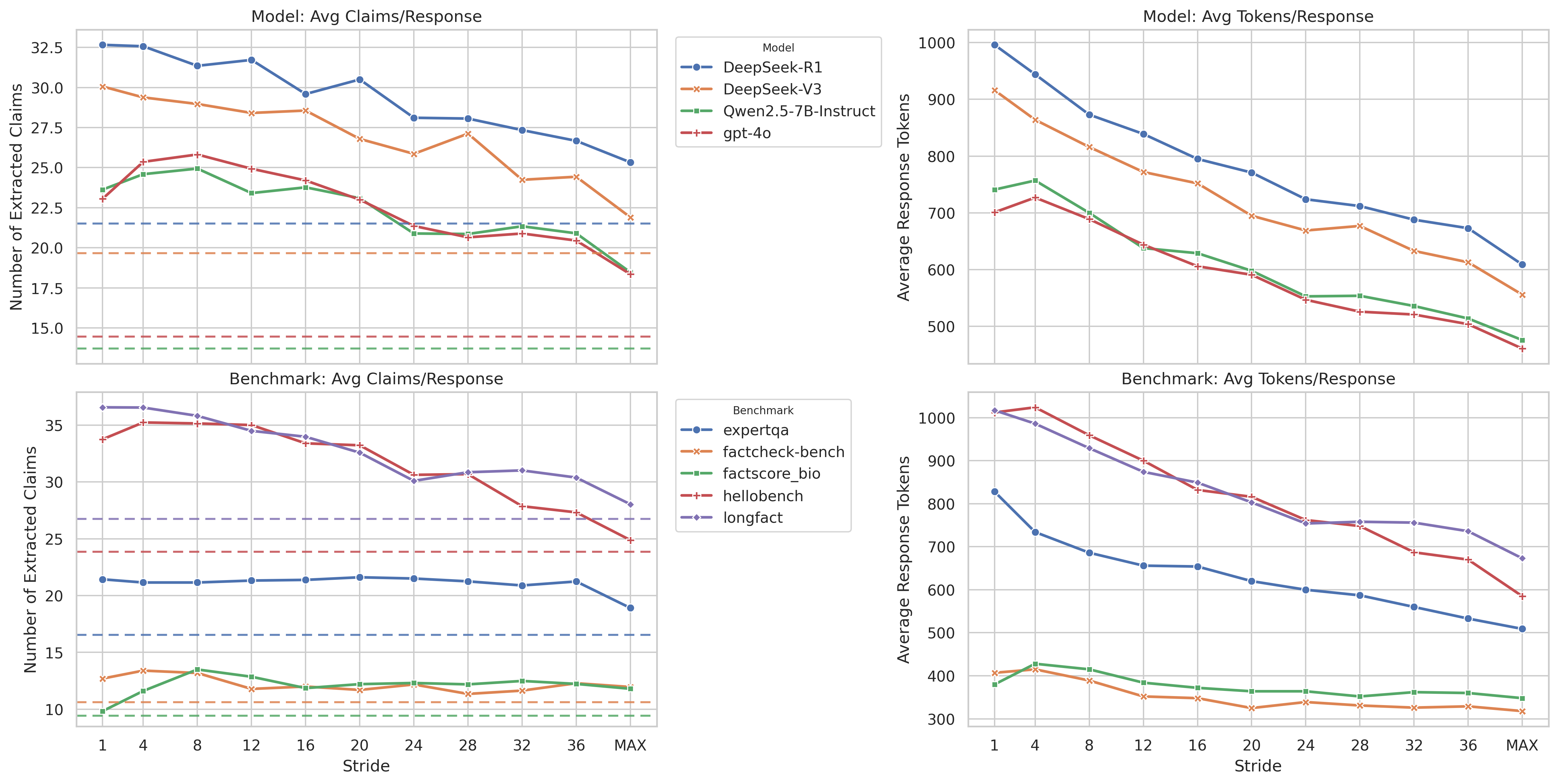}
    \vspace{-5mm}
    \caption{Ablations of altering chunk stride from 1 (= sentence-level window) to MAX (= entire response). 
    The left panel shows the changes in claim extraction \textbf{reliability}, i.e., the discrepancy between \name's number of extracted claims (in solid lines) against ground-truth values (in horizontal dotted lines); the right panel shows the changes in \textbf{efficiency} i.e., token cost.}
    \vspace{-2mm}
\label{fig:ablation-chunk}
\end{figure*}

\subsection{Main Results: \name vs. Baselines}
\label{sec:baselines}

Previous evaluation pipelines often focus on their own development and results, lacking a horizontal comparison with baselines \citep{SAFE_wei2024, veriscore, ExpertQA2024, factool_chern2023}. By utilizing the ground-truth annotations from \name-Bench, we run \name and several baselines in parallel using the same underlying LLM as the extractor and verifier to compare their (1) evaluation reliability, proxied by the alignment with human judgment \textbf{average $|\Delta K|$ and $|\Delta$\metric$|$}, and (2) practical efficiency, proxied by \textbf{total token cost per sample}. From \autoref{tab:baseline-compare}, \name shows significant strength over other baselines, assessed from the perspectives of effectiveness and efficiency, reporting the closest distance to ground-truth \metric score and the ground-truth number of extracted claims.

\vspace{-1mm}

\subsection{Ablations}
\label{sec:ablations}



We analyze the effect of altering chunk stride $\mathbf{w}$ on the extracted claims and the token cost. As observed from~\autoref{fig:ablation-chunk}, \name is decently robust to shorter chunk windows in terms of efficiency and human-machine alignment.

Another non-negligible finding is that as we expand the stride with increasing efficiency, the claim extraction performance does not deteriorate. Rather, the value of strides $\mathbf{w}$ aligned most closely with the ground-truth number of claims ranges from 28 to MAX. This further underscores the over-decomposition issues in previous sentence-level extraction-based evaluations and showcases the effectiveness of using longer chunks as extractor context. It is not a tradeoff between efficiency and accurate claim extraction. Instead, chunk-level decomposition achieves the best of both worlds.

\begin{figure}[htbp]
  \centering
    \includegraphics[width=\linewidth]{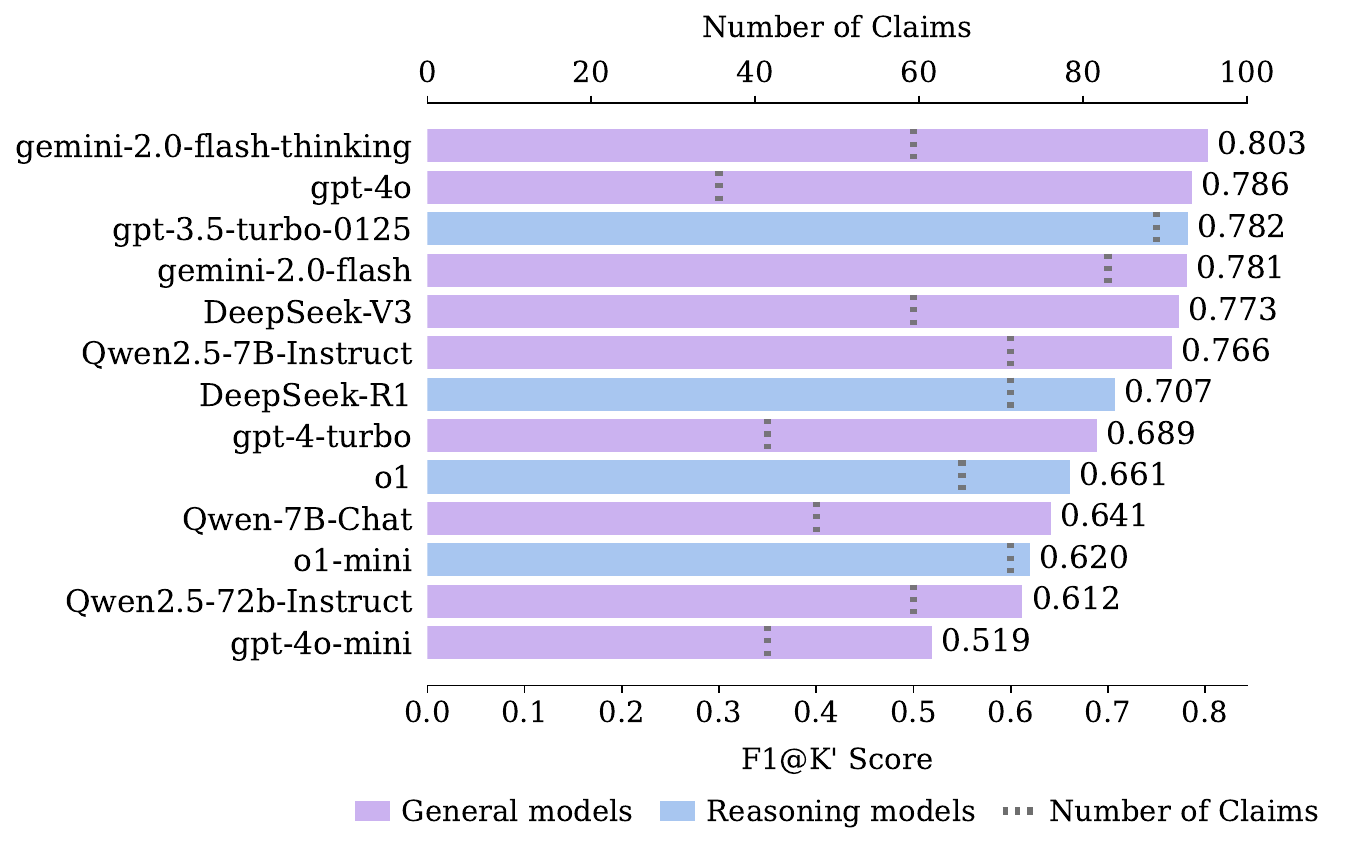}
    \vspace{-15pt}
    \caption{Factuality leaderboard on \name-Bench using \name. The dotted line illustrates the average number of claims decomposed by \name.}
    \label{fig:leaderboard}
    \vspace{-15pt}
\end{figure}

\subsection{Long-Form Factuality Leaderboard Using \name}
\label{sec:model comparison}

After verifying its effectiveness and efficiency, we run \name over state-of-the-art LLMs to investigate how they perform in long-form generation tasks in terms of factuality (\autoref{fig:leaderboard}). Among the LLMs being evaluated, gemini-2.0-flash-thinking tops the \name-Bench evaluation, demonstrating strong performance across sub-domains. The comprehensive results of all models are listed in \autoref{appdx:model-eval}, including efficiency, sub-processes, and final \metric scores.

The long-form factuality leaderboard also yields interesting findings, such as intra-family gaps: Gemini-flash-thinking surpasses Gemini-flash, and Qwen2.5-7B-Instruct even outperforms the much larger Qwen2.5-72B. These results highlight that models' long-form generation factuality is not strictly correlated with model scale, but also reflects model alignment and design choices.
\section{Conclusion}
\label{sec:conclusion}

This paper analyzes the challenges in long-form factuality evaluation, and introduces \name as an attested more robust and scalable solution together with a fully annotated benchmark \name-Bench. In order to tackle issues including pipeline inefficiency, inauthentic claim sets, and insufficient evidence retrieval, \name incorporates several key features: efficient and flexible chunk-based claim extraction, confidence-based pre-verification, and comprehensive document-level evidence retrieval. Future studies could benefit from using \name for fast and reliable factuality evaluation of long-form generations by the continuously advancing models.



\section*{Limitations}
\label{sec:limitations}

Despite its advancements in efficiency and effectiveness, \name has scope limitations. Firstly, the effectiveness of collecting document-level evidence via web-scraping relies heavily on the quality and accessibility of information on the web. If relevant information is scarce, paywalled, or buried deep within less reputable sources, the system's ability to retrieve comprehensive and accurate evidence will be hampered, directly impacting the veracity of the verification process. Furthermore, the empirical results presented, while extensive, are inherently limited by the datasets or domains of long-form LLM generations used in the experiments. We encourage the community to adopt and build upon \name\ for evaluating long-form factuality across more diverse domains and tasks.

\bibliography{main}

\onecolumn
\appendix

\definecolor{light-gray}{gray}{0.95}
\definecolor{boxedge}{gray}{0.85}
\definecolor{boxtitlebg}{gray}{0.90}

\newtcolorbox{comparisonbox}[2][]{
    colback=light-gray,
    colframe=boxedge,
    coltitle=black,
    colbacktitle=boxtitlebg,
    fonttitle=\bfseries\scriptsize, 
    title=#2,
    sharp corners,
    boxrule=0.5pt,
    #1
}

\section{Problem Scope of LLM Factuality}
\label{appdx:problem-scope}

LLM factuality has been a fuzzy notion subject potentially encompassing multiple tasks and model behavior in different discourses. \citet{hallulens} distinguishes between hallucination and factuality, with the former referring to describing the alignment between the model output and the knowledge that the model had access to, either in its training data or as input at inference time, while factuality refers to the alignment between the model output and the objective world knowledge.

Hence, we clarify the problem scope of \textit{factuality evaluation} in this work to fall into testing the absolute factual correctness of the long-form generations, i.e., whether the content of the model response adheres to the information form established knowledge sources (rather than information in the question input or model training data). Specifically, we focus on comparing with the line of research under \citet{factscore}'s decompose-then-verify framework (as listed in related works). This differs from two categories of evaluation works which typically adopt similar terms such as hallucination and facts: (1) factual consistency evaluations on whether inference output aligns with information within the question input, typically applicable to long-context understanding tasks or summarization \citep{RAGAs, ASQA, fabbri-etal-2022-qafacteval}
; (2) fact-checking studies focusing exclusively on classifying claim-level factual correctness \citep{tacl/AtanasovaSLA22, ZhangG23_fact_verification, AVeriTeC, guan_fact_verification_naacl2024, minicheck_emnlp/TangLD24, ZhangG024, FIRE_factchecking, AVerImaTeC}.

\section{Additional Related Work}
\label{appdx:additional_related}

\paragraph{Benchmarking LLM Factuality.}
LLM factuality benchmarks (i.e., evaluating the factual correctness of model output against evidence sourced from established world knowledge \citep{HuCLGWYG24,hallulens}) vary in terms of output length and evaluation granularity. Short-form factuality benchmarks such as SimpleQA \citep{SimpleQA}, TruthfulQA \citep{TruthfulQA}, and HalluQA \citep{HalluQA2023} evaluate the factual correctness of a single factoid directly against groundtruth answers via automated metrics (e.g., BLEURT, ROUGE, and BLEU) or LLM classifiers. Extending to multiple factoids in an LLM generation, early benchmarks annotate LLM factuality on the sentence level \citep{FELM2023, FreshQA2024} or response level \citep{HaluEval2023}. In terms of benchmarking long-form factuality, existing works construct questions from three sources: (1) manually curate questions across expert domains (e.g., medicine, law) \citep{ExpertQA2024}; (2) extract from online discourse platforms \citep{factcheck-bench}, established knowledge sources like Wikipedia \citep{factscore}, or literary corpora \citep{veriscore}; (3) synthesize prompts via LLMs accompanied with explicit instruction templates to elicit longer output \citep{SAFE_wei2024}.

\paragraph{Quantifying Long-Form Factuality in LLM Generations}
Significant strides have been made in long-form
text generation by using LLMs~\citep{TanGSXLFLWSLS24, QiXGWZX24}.
Under the decompose-then-verify framework, the primary quantifying proxy for reporting long-form factuality is factual precision \citep{factscore, DhuliawalaKXRLC24, TianMYMF24}, calculated as the proportion of supported factual claims among total extracted claims. SAFE \citep{SAFE_wei2024} emphasizes on including the coverage (i.e., factual density or informativeness) of the response by quantifying recall and incorporates it into a new F1@K metric, where a score penalty is given towards responses that do not contain a sufficient number of supported claims compared with a domain-averaged number of the automated model evaluations treated as ground-truth.

\section{Claim Extraction Failures in Existing Long-Form Factuality Evaluation Systems}
\label{appdx:parent-claim-extraction-failure-analyses}

\subsection{Failure Cases of Claim Extraction in SAFE}
\label{appdx:failure-cases-SAFE}

The table below demonstrates examples of claim extraction failures under the types and subcategories introduced in \autoref{sec:analysis}. During case inspections, two observations are noteworthy: 

Firstly, \textbf{different sub-categories of claim extraction failures tend to co-occur}. For example, a \textit{subjective} claim is usually \textit{ambiguous}, as subjectivity frequently comes from lacked clarity of a descriptive word or phrase. For another example, over-decomposition of a sentence usually affects all extracted products, inducing both redundant claims and ambiguous claims. 

Secondly, the sentence-level extraction window in existing evaluation pipelines like SAFE causes \textbf{a systematic failure pattern where factual claims composed of fragmented information from multiple sentences are left uncovered} (e.g., something to be summarized from scattered points across several sentences or paragraphs). For instance, Claim 14 in Table \ref{tab: claim_extraction_failures} is a non-existent ground-truth claim that should be extracted from the response by gathering the sequential step information about E-commerce product moderation, but it is missing in SAFE's extraction.


\scalebox{0.9}{
\begin{minipage}{\linewidth}

\begin{longtable}[c]{@{} 
>{\columncolor[HTML]{FFFFFF}\color[HTML]{333333}\RaggedRight\arraybackslash}p{0.35\textwidth} 
>{\columncolor[HTML]{FFFFFF}\color[HTML]{333333}\RaggedRight\arraybackslash}p{0.05\textwidth} 
>{\columncolor[HTML]{FFFFFF}\color[HTML]{333333}\RaggedRight\arraybackslash}p{0.30\textwidth} 
>{\columncolor[HTML]{FFFFFF}\color[HTML]{333333}\RaggedRight\arraybackslash}p{0.15\textwidth} 
>{\columncolor[HTML]{FFFFFF}\color[HTML]{333333}\RaggedRight\arraybackslash}p{0.15\textwidth} 
@{}}

\label{tab: claim_extraction_failures}\\ 

\toprule
{\color[HTML]{333333} Sentence(s)} & 
{\color[HTML]{333333} Claim ID} & 
{\color[HTML]{333333} Claims} & 
{\color[HTML]{333333} Failure Type} & 
{\color[HTML]{333333} Subcategories} \\* 
\midrule
\endfirsthead

\multicolumn{5}{c}%
{{\bfseries Table \thetable\ continued from previous page}} \\
\toprule
{\color[HTML]{333333} Sentence(s)} & 
{\color[HTML]{333333} Claim ID} & 
{\color[HTML]{333333} Claim} & 
{\color[HTML]{333333} Failure Type} & 
{\color[HTML]{333333} Subcategories} \\* 
\midrule
\endhead

\bottomrule
\endfoot

\bottomrule
\endlastfoot

%
\cellcolor[HTML]{FFFFFF}{\color[HTML]{333333} } & 
{\color[HTML]{333333} 1} & 
{\color[HTML]{333333} There are two types of knowledge.} & 
{\color[HTML]{333333} unverifiable, redundant} & 
{\color[HTML]{333333} ambiguous} \\

\cellcolor[HTML]{FFFFFF}{\color[HTML]{333333} } & 
{\color[HTML]{333333} 2} & 
{\color[HTML]{333333} The differences between the two types of knowledge have implications.} & 
{\color[HTML]{333333} unverifiable} & 
{\color[HTML]{333333} ambiguous} \\

\multirow{-8}{=}{%
\color[HTML]{333333} The differences between the two types of knowledge have significant implications in the field of epistemology.%
}  & 
{\color[HTML]{333333} 3} & 
{\color[HTML]{333333} The implications are significant.} & 
{\color[HTML]{333333} unverifiable} & 
{\color[HTML]{333333} ambiguous / subjective} \\

\midrule 
\cellcolor[HTML]{FFFFFF}{\color[HTML]{333333} } &
{\color[HTML]{333333} 4} &
{\color[HTML]{333333} A priori knowledge is justified using reason.} & 
{\color[HTML]{333333} redundant} &
{\color[HTML]{333333} intra- + inter-sentence} \\

\multirow[t]{-3}{=}{\cellcolor[HTML]{FFFFFF}{\color[HTML]{333333}\RaggedRight A priori knowledge is justified independently of experience, using reason and logic.}} &
{\color[HTML]{333333} 5} &
{\color[HTML]{333333} A priori knowledge is justified using logic.} & 
{\color[HTML]{333333} redundant} &
{\color[HTML]{333333} intra- + inter-sentence} \\
\midrule
\cellcolor[HTML]{FFFFFF}{\color[HTML]{333333} } &
{\color[HTML]{333333} 6} &
{\color[HTML]{333333} A posteriori knowledge relies on empirical evidence.} &
{\color[HTML]{333333} redundant} &
{\color[HTML]{333333} intra-sentence} \\
\multirow[t]{-3}{=}{\cellcolor[HTML]{FFFFFF}{\color[HTML]{333333}\RaggedRight On the other hand, a posteriori knowledge relies on empirical evidence for justification.}} &
{\color[HTML]{333333} 7} &
{\color[HTML]{333333} A posteriori knowledge relies on empirical evidence for justification.} &
{\color[HTML]{333333} redundant} &
{\color[HTML]{333333} intra-sentence} \\
\midrule
\cellcolor[HTML]{FFFFFF}{\color[HTML]{333333} } &
{\color[HTML]{333333} 8} &
{\color[HTML]{333333} A priori knowledge is based on pure reason.} & 
{\color[HTML]{333333} redundant} &
{\color[HTML]{333333} intra- + inter-sentence} \\
\multirow[t]{-3}{=}{\cellcolor[HTML]{FFFFFF}{\color[HTML]{333333}\RaggedRight A priori knowledge is often considered to be more certain and necessary, as it is based on pure reason and logic.}} &
{\color[HTML]{333333} 9} &
{\color[HTML]{333333} A priori knowledge is based on logic.} & 
{\color[HTML]{333333} redundant} &
{\color[HTML]{333333} intra- + inter-sentence} \\
\midrule
\cellcolor[HTML]{FFFFFF}{\color[HTML]{333333} } &
{\color[HTML]{333333} 10} &
{\color[HTML]{333333} "2+2=4" is considered to be necessarily true.} &
{\color[HTML]{333333} redundant} &
{\color[HTML]{333333} intra-sentence} \\
\cellcolor[HTML]{FFFFFF}{\color[HTML]{333333} } &
{\color[HTML]{333333} 11} &
{\color[HTML]{333333} "2+2=4" is a mathematical truth.} &
{\color[HTML]{333333} redundant} &
{\color[HTML]{333333} intra-sentence} \\
\multirow[t]{-5}{=}{\cellcolor[HTML]{FFFFFF}{\color[HTML]{333333}\RaggedRight For example, mathematical truths like "2+2=4" are known a priori and are considered to be necessarily true.}} &
{\color[HTML]{333333} 12} &
{\color[HTML]{333333} A truth are considered to be necessarily true.} &
{\color[HTML]{333333} unverifiable} &
{\color[HTML]{333333} tautology} \\
\midrule
{\color[HTML]{333333} A priori knowledge refers to knowledge that is independent of experience, while a posteriori knowledge is knowledge that is derived from experience.} &
{\color[HTML]{333333} 13} &
{\color[HTML]{333333} A priori knowledge refers to knowledge that is independent of experience.} &
{\color[HTML]{333333} missing} &
{\color[HTML]{333333} removed by irrelevance check} \\
\midrule
{\color[HTML]{333333} (E-commerce Product Moderation: \newline Step 1: Automated Content Screening: [...more sentences...] \newline Step 2: Contextual Human Review: [...more sentences...] \newline [...more sentences...])} &
{\color[HTML]{333333} 14} &
{\color[HTML]{333333} (E-commerce Product Moderation contains two steps of automated content screening and contextualized human review.)} &
{\color[HTML]{333333} missing} &
{\color[HTML]{333333} unextracted inter-sentence claims} \\
\caption{Examples of claim extraction failures, inspected from SAFE.}
\end{longtable}
\vspace{0.5em}
\end{minipage}

}

\subsection{Distribution Analysis of Failure Types in VeriScore}
\label{appdx:failure-distribution-analysis-of-veriscore}
In the preliminary analysis of claim extraction in VeriScore, we use Gemini 2.0 Flash (\texttt{gemini-2.0-flash-002}) \citep{kampf2025experiment} in a few-shot LLM-as-judge setting by providing it with the question, response, and claims extracted by VeriScore \citep{veriscore}, to identify specific types of potential extraction failures and their subcategories. The questions and sampled model responses are from a 25\% subset of \name-Bench. The implementation approach follows \citet{Decomposition_Dilemmas}'s analysis on claim decomposition for fact-checking.

\begin{figure}[t]
    \centering
    \includegraphics[width=0.9\linewidth]{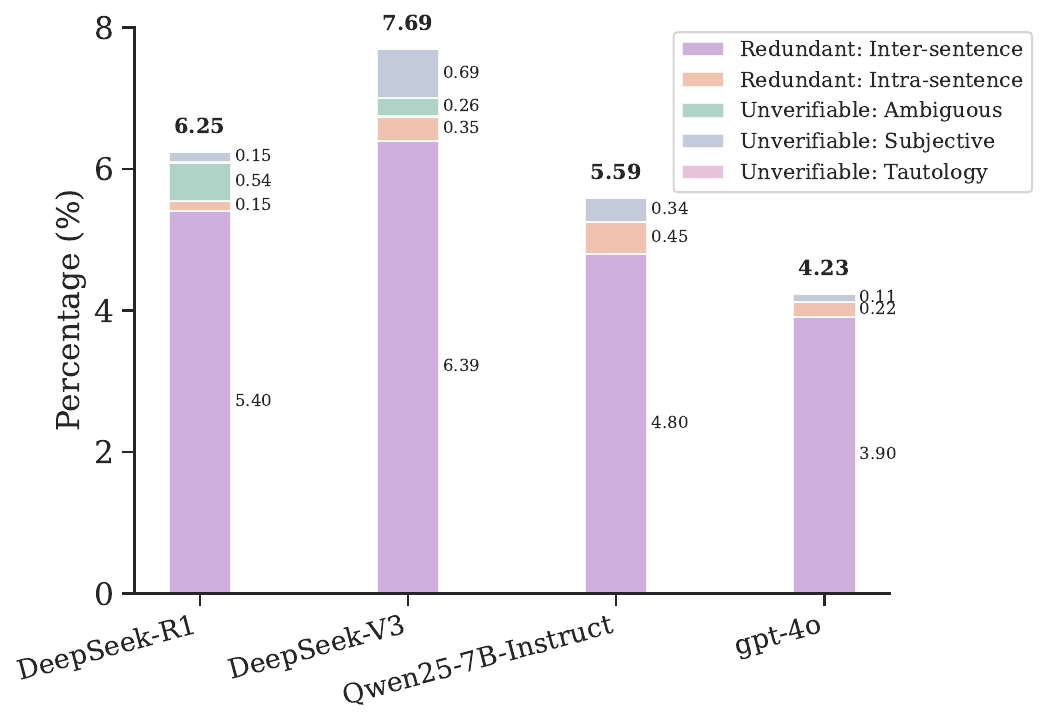}
    \caption{Distribution of unverifiable and redundant claims as well as their subcategories, extracted by VeriScore from sampled model generations. Redundant claims include the labels: `Inter-sentence' and `Intra-sentence'; Unverifiable claims include the labels: `Sujective', `Ambiguous', and `Tautology'. }
    \label{fig:failure-distribution-unverifiable-redundent-claims}
\end{figure}%

Notably, we cannot directly determine whether the extracted sub-claims include missing ones. To address this, we separately prompt the judge to re-examine the generated response and the current set of sub-claims to identify additional missing claims. The missing ratio is then computed as the proportion of missing claims relative to the number of extracted sub-claims (Table~\ref{tab:failure-distribution-missing-claims}). The exact prompt templates are provided in Appendix~\ref{appdx:prompt-error}.

\begin{table}[h]
\centering
\begin{adjustbox}{max width=\linewidth}
\begin{tabular}{lccc}
\toprule
\textbf{Model} & \textbf{Total Missing Claims} & \textbf{Removed by Relevance} & \textbf{Missed by Extractor} \\
\midrule
DeepSeek-R1          & 9.52\%   & 6.98\%   & 2.54\%  \\
DeepSeek-V3          & 11.53\%  & 6.55\%   & 4.98\%  \\
GPT-4o               & 18.68\%  & 11.61\%  & 7.07\%  \\
Qwen2.5-7B           & 15.35\%  & 12.67\%  & 2.67\%  \\
\bottomrule
\end{tabular}
\end{adjustbox}
\caption{Percentage of missing claims and their subcategories, extracted by VeriScore from sampled model generations.}
\label{tab:failure-distribution-missing-claims}
\end{table}

\section{Evidence Search: Case Studies of VeriScore vs. \name{}}
\label{appdx:cases-evidence-search}
This section presents a case study comparing snippet-based and full webpage evidence retrieval approaches.
The example claim for evidence search and verification: is: "The global smartphone market grew 27\% annually from 2010 to 2015."

As shown below, existing evaluation pipelines (e.g., VeriScore, SAFE, FacTool) retrieve short snippets from the Google Serper search engine as evidence. In the example, the retrieved snippets provide fragmented growth figures (5\%-17.9\%) but lack temporal information, let alone offering the complete 2010-2015 data that should be ideally obtained to verify the 27\% annual growth claim. The limited information in the evidence snippets renders the search quality poor and prevents meaningful verification in the subsequent step.

\begin{tcolorbox}[
    breakable,
    enhanced,
    colback=blue!4!white,
    colframe=black!5!white,
    title=VeriScore: Snippet-based Evidence,
    fonttitle=\bfseries,
    coltitle=black,
    boxrule=1pt,
    arc=3mm,
    drop fuzzy shadow,
    before upper={\parindent0pt\small}
]

\begin{tabularx}{\linewidth}{@{}>{\RaggedRight}p{0.25\linewidth}>{\RaggedRight}p{0.55\linewidth}>{\RaggedRight\arraybackslash}p{0.2\linewidth}@{}}
\toprule
\textbf{Title} & \textbf{Snippet} & \textbf{Link} \\
\midrule

Smartphone Market Share - IDC & 
``The U.S. smartphone market experienced growth of more than 5\%, despite challenges from tariffs and trade wars affecting disposable income,'' \ldots & 
\href{https://www.idc.com/promo/smartphone-market-share/}{IDC Market Share} \\

\addlinespace

Smartphone Market Insights - IDC & 
Detailed market size and share trends empower companies selling Mobile Phones to get ahead of market changes and compete more effectively. & 
\href{https://www.idc.com/promo/smartphone-market-share/os}{IDC Insights} \\

\addlinespace

Apple Grabs Top Spot 2023 & 
``The fourth quarter (4Q23) saw 8.5\% year-over-year growth and 326.1 million shipments, higher than the forecast of 7.3\% growth.'' & 
\href{https://my.idc.com/getdoc.jsp?containerId=prUS51776424}{IDC Report} \\

\addlinespace

Worldwide Smartphone Market 2015 & 
``IDC predicts worldwide smartphone shipments will grow 9.8\% in 2015 to a total of 1.43 billion units, the first full year of single-digit \ldots'' & 
\href{https://www.businesswire.com/news/home/20151203005661/en/}{BusinessWire} \\

\addlinespace

Cisco Annual Internet Report & 
The mobile M2M category is projected to grow at a 30\% CAGR from 2018 to 2023. Smartphones will grow at a 7\% CAGR within the same period. & 
\href{https://www.cisco.com/c/en/us/solutions/collateral/}{Cisco Report} \\

\addlinespace

Statista Sales 2007-2023 & 
``In 2022, smartphone vendors sold around 1.39 billion smartphones worldwide, with this number forecast to drop to 1.34 billion in 2023.'' & 
\href{https://www.statista.com/statistics/263437/}{Statista} \\

\addlinespace

Smartphone - Wikipedia & 
``Since the early 2010s, improved hardware and faster wireless communication have bolstered the growth of the smartphone industry.'' & 
\href{https://en.wikipedia.org/wiki/Smartphone}{Wikipedia} \\

\addlinespace

Smartphones (Global Market) & 
``Smartphone sales grew by 4\% over the year - data from Counterpoint Technology Market Research, Samsung leads the \ldots'' & 
\href{https://tadviser.com/index.php/Article:Smartphones}{TAdviser} \\

\addlinespace

Mobile Phone Market 2010 & 
``The worldwide mobile phone market grew 17.9\% in 4Q10 on shipments of 401.4 million units (IDC Worldwide Quarterly Mobile).'' & 
\href{https://www.businesswire.com/news/home/20110127006929/en/}{BusinessWire} \\

\addlinespace

Best-Selling Phones & 
``Global Annual Smartphone Market Grew for the First Time Since 2017; Shipments Suffer Largest-Ever Decline with 18.3\% Drop.'' & 
\href{https://en.wikipedia.org/wiki/List_of_best-selling_mobile_phones}{Wikipedia} \\

\bottomrule
\end{tabularx}

\begin{tcolorbox}[
    colback=blue!2!white,
    colframe=blue!30!white,
    title=Verification Label,
    fonttitle=\bfseries\small,
    coltitle=black,
    boxrule=0.5pt,
    arc=2mm,
    left=2mm,right=2mm,
    top=1mm,bottom=1mm
]
\texttt{Not Enough Evidence} \textcolor{red}{\textbf{\ding{55}\ [GroundTruth Label: Refuted]}} \\
\end{tcolorbox}

\end{tcolorbox}

In contrast, below for the same claim, \name retrieves full webpage content as document-level knowledge sources and then retrieves relevant chunks from these documents for the verifier, as demonstrated below. The comprehensive retrieved evidence provides informative context for the verifier to reason over whether the target claim can be supported, empowering traceable and accurate claim verification results, effectively boosting reliability and transparency.

\begin{tcolorbox}[
    breakable,
    enhanced,
    colback=blue!4!white,
    colframe=black!5!white,
    title=FastFact: Scraped Evidence Documents from Full Webpages,
    fonttitle=\bfseries,
    coltitle=black,
    boxrule=1pt,
    arc=3mm,
    drop fuzzy shadow,
    before upper={\parindent0pt} 
]

\begin{description}[
    leftmargin=!,
    labelwidth=0pt,
    style=nextline,
    font=\normalfont,
    itemsep=10pt, 
    topsep=5pt
]
    \item[\textbf{Evidence 1}]

    \textbf{Source Title:} Smartphone Sales Statistics By Revenue, Samsung Sales, Apple Sales and Facts

    \textbf{Content:} people in the U.S. were using Android devices. As of January 2023, about 25\% of Android phones worldwide were running version 12 of Google's operating system. By February 2023, 81\% of Apple devices using the App Store had iOS 16 installed, while 15\% were still using iOS 15. 
    
    \begin{center}
    Global Smartphone Sales Statistics (Reference: statista.com)
        
        \begin{tabular}{l S[table-format=4.2]} 
        \toprule
        \textbf{Year} & {\textbf{Sales (million units)}} \\ 
        \midrule
        2007 & 122.32 \\
        2008 & 139.29 \\
        2009 & 172.38 \\
        2010 & 296.65 \\
        2011 & 472.00 \\ 
        2012 & 680.11 \\
        2013 & 969.72 \\
        2014 & 1244.74 \\
        2015 & 1423.90 \\ 
        2016 & 1495.96 \\
        2017 & 1536.54 \\
        2018 & 1556.27 \\
        2019 & 1540.66 \\
        2020 & 1351.84 \\
        2021 & 1433.86 \\
        2022 & 1395.25 \\
        2023 & 1339.51 \\
        \bottomrule
        \end{tabular}
    \end{center}
        
    In 2022, about 1.40 billion mobile phones were sold globally, but this number is expected to drop slightly to 1.36 billion in 2024. In 2016, less than half of the world's population owned a smartphone. However, the percentage of people with smartphones has been rising steadily, reaching 78.05\% in 2020. By 2025, nearly 87\% of mobile users in the U.S. are expected to own a smartphone, up from just 27\% in 2010. Smartphone sales in the U.S. were projected to reach around \$73 billion in 2021, compared to \$18 billion in 2010.

    \item[\textbf{Evidence 2}]
    \textbf{Source Title:} Smartphone Sales Statistics By Revenue, Samsung Sales, Apple Sales and Facts

    \textbf{Content:} just 27\% in 2010. Smartphone sales in the U.S. were projected to reach around \$73 billion in 2021, compared to \$18 billion in 2010. Global smartphone sales are also expected to grow in 2021 across all regions as the market recovers from the initial impact of the COVID-19 pandemic. Global Shipment Statistics In 2023, smartphone performance varied by region. The Middle East and Africa (MEA) saw strong growth in shipments, while North America experienced the largest decline. Among the top 10 smartphone brands, Motorola, Tecno, Huawei, and HONOR saw double-digit growth compared to the previous year. Infinix and Itel, which are sister brands to Tecno, also had double-digit growth. (Source: sci-tech-today.com) In the first quarter of 2024, Samsung was the top smartphone maker worldwide, shipping around 60 million units, which gave it 21\% of the market share. Apple came in second place, shipping about 50 million units, down from 55 million in the same quarter of the previous year, according to Smartphone Sales Statistics. Apple held 17\% of the market. During this period, nearly 290 million units of smartphones were shipped.

    \item[\textbf{Evidence 3}]
    \textbf{Source Title:} Smartphone Sales Statistics By Revenue, Samsung Sales, Apple Sales and Facts
    
    \textbf{Content:} 2024. In 2016, less than half of the world's population owned a smartphone. However, the percentage of people with smartphones has been rising steadily, reaching 78.05\% in 2020. By 2025, nearly 87\% of mobile users in the U.S. are expected to own a smartphone, up from just 27\% in 2010. Smartphone sales in the U.S. were projected to reach around \$73 billion in 2021, compared to \$18 billion in 2010. Global smartphone sales are also expected to grow in 2021 across all regions as the market recovers from the initial impact of the COVID-19 pandemic. Global Shipment Statistics 
    In 2023, smartphone performance varied by region. The Middle East and Africa (MEA) saw strong growth in shipments, while North America experienced the largest decline. Among the top 10 smartphone brands, Motorola, Tecno, Huawei, and HONOR saw double-digit growth compared to the previous year. Infinix and Itel, which are sister brands to Tecno, also had double-digit growth. (Source: sci-tech-today.com) 
    In the first quarter of 2024, Samsung was the top smartphone maker worldwide, shipping around 60 million units, which gave it 21\% of the market share.

    \item[\textbf{Evidence 4}]
    \textbf{Source Title:} Smartphone Sales Statistics By Revenue, Samsung Sales, Apple Sales and Facts
    
    \textbf{Content:} use today, and this number is expected to keep rising. The global smartphone market grew by 7.8\% in the first quarter of 2024 alone. The smartphone market is worth billions of dollars, and companies are competing to sell you their phones. We’ve put together all the key information about this battle in our Smartphone Sales Statistics collection. Editors Choice In the first quarter of 2023, global smartphone shipments totaled 280.2 million units, a 14\% decrease from the same period last year. During this time, Samsung was the top-selling smartphone brand, helped by the launch of its Galaxy S23 series. By mid-2023, there were 6.92 billion smartphone users worldwide. Young adults aged 18 to 49 are the biggest users of smartphones. According to Smartphone Sales Statistics, smartphone manufacturers shipped 43 billion units globally in 2023, down 4\% from the previous year. According to International Data Corporation (IDC), smartphone shipments worldwide reached 34 million units in the first half of 2023, a drop from the same period last year. The top-selling smartphone worldwide is the iPhone 14 Pro Max, with 26.5 million units sold. In recent years, around 5 billion smartphones have been sold each year across the globe.

    \item[\textbf{Evidence 5}]
    \textbf{Source Title:} Smartphone Sales Statistics By Revenue, Samsung Sales, Apple Sales and Facts
    
    \textbf{Content:} Other brands: 29\% For the second half of 2023, the market share distribution was: Samsung: 20\% Apple: 17\% Xiaomi: 12\% Oppo: 10\% Vivo: 8\% Other brands: 33\%. Of the first five mobile brands, Xiaomi was just the one to show year-over-year development; it shipped nearly 42.5 million mobile phones in 2023. Samsung’s revenue fell by 8\% year-over-year in 2023, even though it grew by 11\% from the previous quarter. Despite this, Samsung remained the leading smartphone brand. Overall, the global smartphone market shrank by 1\% year-over-year, reaching a total of 299.8 million units shipped in 2023.

    \item[\textbf{Evidence 6}]
        \textbf{Source Title:} Smartphone Sales Statistics By Revenue, Samsung Sales, Apple Sales and Facts
        
        \textbf{Content:} Revenue Statistics iPhone vs Android Smartphone Sales in Different Nations Xiaomi Smartphone Sales Statistics Conclusion Introduction Smartphone Sales Statistics: Smartphones are very common in developed countries. In fact, nearly 75\% of people in the top 10 developed countries own one. Around 90\% of mobile phones worldwide are smartphones, and most people worldwide have one. There are more than 7.2 billion smartphones in use today, and this number is expected to keep rising. The global smartphone market grew by 7.8\% in the first quarter of 2024 alone. The smartphone market is worth billions of dollars, and companies are competing to sell you their phones. We've put together all the key information about this battle in our Smartphone Sales Statistics collection. Editors Choice
        In the first quarter of 2023, global smartphone shipments totaled 280.2 million units, a 14\% decrease from the same period last year. During this time, Samsung was the top-selling smartphone brand, helped by the launch of its Galaxy S23 series. By mid-2023, there were 6.92 billion smartphone users worldwide. Young adults aged 18 to 49 are the biggest users of smartphones.
    
    \item[\textbf{Evidence 7}]
        \textbf{Source Title:} Competition, Product Proliferation, and Welfare: A Study of the US Smartphone Market
        
        \textbf{Content:} of the US smartphone market. The smartphone industry has been one of the fastest growing industries in the world, with billions of dollars at stake. Worldwide smartphone sales grew from 122 million units in 2007 to 1.4 billion units in 2015 (Gartner, Inc. 2007, 2015), with about \$400 billion in global revenue in 2015 (Nuremberg 2016). Moreover, product proliferation is a prominent feature of this industry. For example, in the US market during our sample period, Samsung, on average, simultaneously offered 11 smartphones with substantial quality and price variation. In order to address our research questions, we develop a structural model of consumer demand and firms' product and pricing decisions. We estimate the model using data from the Investment Technology Group (ITG) Market Research. This
        dataset provides information on all smartphone products in the US market between
        January 2009 and March 2013. For every month during this period, we observe both
        the price and the quantity of each smartphone sold through each of the four national
        carriers in the United States (AT\&T, T-Mobile, Sprint, and Verizon).
    
    \item[\textbf{Evidence 8}]
        \textbf{Source Title:} Competition, Product Proliferation, and Welfare: A Study of the US Smartphone Market
        
        \textbf{Content:} to those products that would remain available, this can constitute a harm to customers over and above any effects on the price or quality of any given product. If there is evidence of such an effect, the Agencies may inquire whether the reduction in variety is largely due to a loss of competitive incentives attributable to the merger." We study our research questions in the context of the US smartphone market. The smartphone industry has been one of the fastest growing industries in the world, with billions of dollars at stake. Worldwide smartphone sales grew from 122 million
        units in 2007 to 1.4 billion units in 2015 (Gartner, Inc. 2007, 2015), with about
        \$400 billion in global revenue in 2015 (Nuremberg 2016). Moreover, product proliferation is a prominent feature of this industry. For example, in the US market during
        our sample period, Samsung, on average, simultaneously offered 11 smartphones
        with substantial quality and price variation. In order to address our research questions, we develop a structural model of
        consumer demand and firms' product and pricing decisions. We estimate the model
        using data from the Investment Technology Group (ITG) Market Research.
    
    \item[\textbf{Evidence 9}]
        \textbf{Source Title:} Smartphone Sales Statistics By Revenue, Samsung Sales, Apple Sales and Facts
        
        \textbf{Content:} quarter of the previous year, according to Smartphone Sales Statistics. Apple held 17\% of the market. During this period, nearly 290 million units of smartphones were shipped. Looking ahead to 2027, the expected shipment numbers by price category are: \$92 million for 4G smartphones (based on average selling price, or ASP) \$377 million for standard premium smartphones (SP ASP) \$433 million for 5G smartphones (5G ASP). As of April 2023, the smartphone industry made the following revenue: \$159 million from 4G smartphones \$584 million from standard premium smartphones \$421 million from 5G smartphones. In the first quarter of 2023, the leading smartphone brands and their market share were: Samsung: 22\% Apple: 21\% Xiaomi: 11\%* Oppo: 10\% Vivo: 7\% Other brands: 29\% For the second half of 2023, the market share distribution was: Samsung: 20\% Apple: 17\% Xiaomi: 12\% Oppo: 10\% Vivo: 8\% Other brands: 33\%. Of the first five mobile brands, Xiaomi was just the one to show year-over-year development; it shipped nearly 42.5 million mobile phones in 2023. Samsung's revenue fell by 8\% year-over-year in 2023, even though it grew by 11\% from the previous quarter. Despite this, Samsung remained the leading smartphone brand.
    
    \item[\textbf{Evidence 10}]
        \textbf{Source Title:} Smartphone Sales Statistics By Revenue, Samsung Sales, Apple Sales and Facts
        
        \textbf{Content:} new models, keeping it competitive. Over the years, the iPhone has become a key product for Apple, and its contribution to the company's total revenue has grown steadily. According to Smartphone Sales Statistics, iPhone sales made up about 25\% of Apple's total income in Q1 2009. By Q3 2024, that share increased to nearly 50\%, bringing in almost \$40 billion in that quarter alone. In terms of sales volume, iPhone sales have grown significantly: from 55.8 million units sold in 2010 to 232 million units in 2023. The highest number of iPhones sold in a single quarter was over 90 million in Q4 2020.
        iPhone vs Android Smartphone Sales in Different Nations
        According to smartphone sales data from Kantar, in Q4 2023, iPhones outsold Android in two countries: the United States and Australia. In the U.S., iPhones made up 55.7\% of all smartphone sales, and in Australia, they accounted for 50.3\% of sales. However, in all other countries included in the study, Android was the most popular choice among people buying new smartphones in Q4 2023.

\end{description}


\begin{tcolorbox}[
    colback=blue!2!white,
    colframe=blue!30!white,
    title=Reasoning for Verification Label,
    fonttitle=\bfseries,
    coltitle=black,
    boxrule=0.5pt,
    arc=2mm,
    left=2mm,right=2mm,
    top=1mm,bottom=1mm
]


    The claim states that the global smartphone market grew 27\% annually from 2010 to 2015. The evidence provided includes various statistics about global smartphone sales over the years, specifically mentioning the number of units sold from 2010 to 2015. 
    
    From the evidence, we can see the following sales figures:
    
    \begin{center}
    \footnotesize
    \begin{tabular}{lc}
    \toprule
    Year & Sales (million units) \\
    \midrule
    2010 & 296.65 \\
    2011 & 472.00 \\
    2012 & 680.11 \\
    2013 & 969.72 \\
    2014 & 1,244.74 \\
    2015 & 1,423.90 \\
    \bottomrule
    \end{tabular}
    \end{center}
    
    To determine the annual growth rate, we can calculate the growth from year to year:
    \[
    \text{GR} = \frac{\text{Sales}_t - \text{Sales}_{t-1}}{\text{Sales}_{t-1}} \times 100\%
    \]
    \begin{center}
    \footnotesize
    \begin{tabular}{lc}
    \toprule
    Period & Growth Rate \\
    \midrule
    2010-11 & 59.1\% \\
    2011-12 & 44.1\% \\
    2012-13 & 42.6\% \\
    2013-14 & 28.4\% \\
    2014-15 & 14.4\% \\
    \bottomrule
    \end{tabular}
    \end{center}
    
    The growth rates calculated do not support the claim of a consistent 27\% annual growth rate from 2010 to 2015. In fact, the growth rates vary significantly, with the highest being 59\% in the first year and decreasing thereafter.

    Therefore, the claim is contradicted by the evidence, as the growth rate does not align with the stated 27\% annual growth.

    The evidence \textit{contradicts} the claim of consistent 27\% annual growth, warranting a \texttt{Refuted} label.

\end{tcolorbox}

\begin{tcolorbox}[
    colback=blue!2!white,
    colframe=blue!30!white,
    title=Verification Label,
    fonttitle=\bfseries\small,
    coltitle=black,
    boxrule=0.5pt,
    arc=2mm,
    left=2mm,right=2mm,
    top=1mm,bottom=1mm
]
\texttt{Refuted \textbf{\textcolor{green}{\ding{51}\ [GroundTruth Label: Refuted]}}}
\end{tcolorbox}

\end{tcolorbox}

\section{Efficiency Analysis of Existing Long-Form Factuality Evaluation Systems}
\label{appdx:further-complexity-analysis}

We first define the key parameters and cost drivers that form the basis of our comparative analysis.

\subsection{Key Parameters}
\begin{itemize}
    \item[$\mathbf{N}$:] Number of input sentences in the generated text to be evaluated.
    \item[$\mathbf{M}$:] Number of atomic claims extracted from the $N$ sentences.
    \item[$\mathbf{q}$:] Custom number of search queries for evidence retrieval for each claim ($\mathbf{q} \ge 1$).
    \item[$\mathbf{p}$:] Ratio of extracted claims in \name whose pre-verification confidence $C$ did not pass the threshold $\theta$, triggering subsequent evidence retrieval and verification ($0 \le \mathbf{p} \le 1$).
    \item[${\mathbf{w}}$:] Custom chunk stride in \name i.e., the number of sentences provided to the LLM extractor for claim extraction ($1 \le \mathbf{w} \le N$). Prior baselines use sentence-level context window, thus ${\mathbf{w}} = 1$.
\end{itemize}

\subsection{Cost Drivers}
The total cost and processing time of an evaluation pipeline is dominated by two operations:
\begin{enumerate}
    \item \textbf{LLM Inference Calls}: The primary cost driver, as each call to a large API model like GPT-4 incurs significant financial cost and latency.
    \item \textbf{Evidence Retrieval}: Calls to external search APIs introduce latency and potential financial cost. Local retrieval is cheaper but may have lower evidence quality.
\end{enumerate}

\subsection{Inference Cost and Time Complexity}
\autoref{apptab:inference_calls} breaks down the minimum number of LLM inference calls required for each pipeline phase.

\begin{table}[h!]
\centering
\footnotesize
\setlength{\tabcolsep}{3pt} 
\begin{tabular*}{\linewidth}{@{\extracolsep{\fill}}lccccc@{}}
\toprule
\textbf{Pipelines} & \textbf{Extractor Calls} & \textbf{Searches} & \textbf{Verifier Calls} & \textbf{Inference Complexity} & \textbf{Search Complexity} \\
\midrule
FactScore \citep{factscore} & $\mathbf{N}$ & $0$ & $\mathbf{M}$ & $O_{LLM}(\mathbf{N} + \mathbf{M})$ & $0$ \\
SAFE \citep{SAFE_wei2024} & $\mathbf{N} + 2\mathbf{M} (+ \mathbf{q}\mathbf{M})$ & $\mathbf{q}\mathbf{M}$ & $\mathbf{M}$ & $O_{LLM}(\mathbf{N} + (3 + \mathbf{q})\mathbf{M})$ & $O_{search}(\mathbf{qM})$ \\
VeriScore \citep{veriscore} & $\mathbf{N}$ & $\mathbf{q}\mathbf{M}$ & $\mathbf{M}$ & $O_{LLM}(\mathbf{N} + \mathbf{M})$ & $O_{search}(\mathbf{qM})$ \\
\name (Ours) & $\mathbf{N} / \mathbf{w}$ & $\mathbf{p}\mathbf{q}\mathbf{M}$ & $\mathbf{p}\mathbf{M}$ & $O_{LLM}(\mathbf{N}/\mathbf{w} + \mathbf{pM})$ & $O_{search}(\mathbf{pqM})$ \\
\bottomrule
\end{tabular*}
\caption{Breakdown of LLM Inference Calls, Search Queries, and Search Complexity by Pipeline Phase.}
\label{apptab:inference_calls}
\end{table}

\begin{itemize}
    \item \textbf{FactSCore} employs a single LLM call for claim extraction ($\mathbf{N}$) and one for verification per claim ($\mathbf{M}$). Its use of a local retriever avoids external search API calls altogether.
    \item \textbf{SAFE} has the highest base inference cost and time complexity. It extracts claims from each of the $\mathbf{N}$ sentences, then performs two additional LLM calls per claim for \textit{revision} and \textit{relevance checking} ($2\mathbf{M}$). Next, SAFE conducts an iterative multi-step search process, where an LLM is tasked to generate a new query sequentially for $\mathbf{q}$ times based on the claim, the combination of previously generated search queries, and all their corresponding searched results. This induces an extra $\mathbf{q}\mathbf{M}$ inference calls (added in the ``Extractor Calls" column) for $\mathbf{q}\mathbf{M}$ searches. Finally, SAFE performs a verification per claim.
    \item \textbf{VeriScore} first extracts claims from each sentence using a sliding window ($\mathbf{N}$ calls), then collects search results using each claim as the query ($\mathbf{q}\mathbf{M}$ queries), and finally verifies each claim individually ($\mathbf{M}$ verified claims). The majority of existing factuality evaluations follow this standard pattern of LLM calls and searches, often with extra costs at the extraction stage for checks and revisions \citep{factcheck-bench, ExpertQA2024, factool_chern2023}.
    \item \textbf{\name} minimizes LLM inference calls and searches in two aspects: its first expands the extractor's context window from sentences to chunks, reducing the extractor calls from $\mathbf{N}$ to ($\mathbf{N}/\mathbf{w}$); meanwhile, the extractor model also performs confidence-based pre-verification to filter the $\mathbf{M}$ claims down to a subset of $\mathbf{p}\mathbf{M}$ for further search and verification. This scales down both search queries and LLM inference calls sufficiently to $\mathbf{p}\mathbf{q}\mathbf{M}$ and $(\mathbf{N}/\mathbf{w} + \mathbf{p}\mathbf{M})$.
\end{itemize}






\section{Benchmark Construction}
\label{sec:benchmark_construction}
\subsection{Construction}
Following the goal, our selection of benchmark questions must meet the following criteria: (1) factually intensive and knowledge-based; (2) eliciting long-form generations across lengths; (3) diverse domains and task types, as listed in Table \ref{tab:fastfact_bench_stats}. Specifically, to obtain a diverse coverage of question types and response structures, we carefully sampled knowledge-based questions from a long-form QA (LFQA) benchmark HelloBench \citep{hellobench} on top of the established four factuality benchmarks. This raises the length of long-form factuality evaluation to an average of 51 sentences, challenging evaluation metrics on significantly longer generations.

To reliably evaluate the effectiveness our evaluation method (i.e., alignment with human judgement), we collect human evaluation groundtruths throughout the decompose-then-verify framework about claim decomposition and verification in \name-Bench. After aggregating 100 long-form factuality questions (20 each from the listed benchmarks), we sample responses from four representative models (DeepSeek-R1, DeepSeek-V3, GPT-4O, Qwen-2.5-Instruct), and run \name on the 4*100 model responses. Lastly, 10 annotators review and annotate the products by \name across the subprocesses in our pipeline, thus constructing the complete \name-Bench.

\subsection{Human Annotation}

\paragraph{Multi-Stage Annotation}
Long-form factuality evaluation is extremely labor-intensive and time-consuming \citep{factcheck-bench, ExpertQA2024, factool_chern2023}. Based on previous failure distribution analyses (\ref{sec:analysis}), we divide the annotation of \name-Bench into two tasks: 
(1) revise ambiguous claims/add missing claims/remove redundant claims, which aims to both provide groundtruths and assess \name's performance during the claim extraction subprocess; 
(2) verify claims based on provided/manually-searched evidence, depending on whether \name's pre-verification label passes the confidence check (i.e., assessing pre-verification success and verification success). 
To ease the annotation process, we designed an annotation tool for revision and labeling. The detailed annotation process is described in the Annotation Protocol (attached in \autoref{appdx:anntation-protocol}), displaying the task screens on the annotation tool interface.
Each annotator is assigned 40 samples, which took around 10-16.5 hours per annotator, accumulating to over 500 hours spent on human annotation.  

\paragraph{Annotation Products and Alignment Metrics}

Differing from prior works with limited inspection on the evaluation pipeline \citep{veriscore, SAFE_wei2024, factool_chern2023}, our multi-stage annotation tasks produce ground-truths on both claim extraction and verification, as well as human-machine alignment rates in \name-Bench, specifically:
\begin{enumerate}
    \item \textit{Modifications on the extracted claims}, including (1) revision for unverifiable claims, (2) addition for missing claims, and (3) removal for redundant claims, shown in Table \ref{tab:annotation_delta}. This is the most fine-grained type of quality evaluation, as the modification serves as a reliable proxy for active human judgment on each claim's content quality and validity in \name{}, as well as the groundtruth list of ideally extracted claims;
    
    \item \textit{Ground-truth number of extracted claims from each sampled generation}, including the number of supported, non-supported, and irrelevant facts (\({S'(y), N(y)', I(y')}\)). The detailed statistics grouped by each benchmark among the aggregated \name-Bench in listed in Appendix~\ref{appdx:additional-bench-stats};
    
    \item \textit{Human-Machine alignment}: The alignment between human judgment and automated evaluation pipelines is a widely adopted proxy for system reliability and effectiveness \citep{factscore, factcheck-bench, veriscore}. With the groundtruths collected above, we aggregate the absolute human-\name variance per sample in both subprocesses of claim extraction and verification, including the number of extracted claims, distributions of final verification label, and final end-to-end evaluation score. Table~\ref{tab:annotation_delta} comprehensively shows a high alignment with human decisions across stages of extraction and verification.
\end{enumerate}

\paragraph{Annotation Quality}
To ensure annotation quality, meta-annotation is conducted on a 10\% subset of the entire benchmark, where annotators cross-annotate the same subset of samples, reporting a final inter-rater agreement rate of 93.6\% in Task 1 (using the operation type classification on each original claim generated by \name as agreement proxy), and 95.0\% in Task 2 (using the verification label classification of the original claims as agreement proxy).

\section{Additional Statistics from \name-Bench}
\label{appdx:additional-bench-stats}
\begin{table}[H]
  \small
  \setlength{\tabcolsep}{4pt}
  \begin{tabularx}{\textwidth}{lYYYYYY}
    \toprule
    \thead{\textbf{Benchmark}\\\textbf{Ground-Truths}} &
    \thead{\textbf{Factcheck-}\\\textbf{bench}} &
    \thead{\textbf{Hello-}\\\textbf{bench}} &
    \textbf{Longfact} &
    \textbf{ExpertQA} &
    \textbf{FactScore} &
    \thead{\textbf{\name-}\\\textbf{Bench}\\\textbf{(Agg)}} \\
    \midrule
    \# extracted claims ($S'\!+\!N'\!+\!I'$) & 10.6  & 23.86 & 26.71 & 16.54 &  9.43 & 17.33 \\
    \midrule
    \% supported ($S'$)                         & 74.49\% & 89.00\% & 85.35\% & 86.30\% & 45.97\% & 76.23\% \\
    \% non-supported ($N'$)                     & 22.52\% & 10.84\% & 14.44\% &  7.59\% & 37.69\% & 18.54\% \\
    \quad\% refuted                           &  8.98\% &  1.71\% &  4.47\% &  1.37\% & 23.95\% &  8.06\% \\
    \quad\% not enough evidence               & 12.63\% &  7.63\% &  8.71\% &  5.19\% & 10.88\% &  8.97\% \\
    \quad\% conflicting evidence              &  0.91\% &  1.50\% &  1.26\% &  1.03\% &  1.75\% &  1.29\% \\
    \quad\% others                          &  0.00\% &  0.00\% &  0.00\% &  0.00\% &  1.11\% &  0.22\% \\
    
    \% irrelevant ($I'$)                        &  0.49\% &  0.16\% &  0.22\% &  0.16\% &  0.09\% &  0.22\% \\
    \midrule
    $K'$ ($= S' + N'$)                        & 10.55   & 23.82   & 26.64   & 16.52   &  9.42   & 17.39   \\
    $F_{1}\!@K'$                 & 80.40\% & 88.20\% & 83.30\% & 87.70\% & 57.00\% & 77.70\% \\
    \bottomrule
  \end{tabularx}
  \caption{Annotated groundtruths of \name{}-Bench provide subprocess-level golden values for baseline metrics to align. "\#" represents the average number of the described claims per sample, while "\%" represents the proportion over all extracted claims $(S' + N' + I')$. "\% others" refers to occasions where the annotators directly agree with the pre-verification label of `non-support' without considering evidence-based verification labels.}
\end{table}

\section{\name Evaluation Statistics on State-of-the-Art LLMs}
\label{appdx:model-eval}
\begin{center}
\setlength\LTleft{0pt} 
\setlength\LTright{0pt} 
\scriptsize
\begin{longtable}[c]{lccccccccccc}
\caption{Model Performance Comparison (ranked by F1 Score). Data automatically filled from CSV file.}
\label{tab:model_eval} \\
\toprule
\textbf{Model} 
  & \textbf{Benchmark} 
  & \textbf{Resp Len.} 
  & \textbf{Tokens} 
  & \textbf{Cost(\$)} 
  & \textbf{\#Sents} 
  & \textbf{\#Claims} 
  & \textbf{\#Support} 
  & \textbf{\%Pre-verify} 
  & \textbf{P} 
  & \textbf{F1@K'} \\
\midrule
\endfirsthead
\multicolumn{11}{c}%
{\bfseries \tablename\ \thetable{} -- continued from previous page} \\
\toprule
\textbf{Model} 
  & \textbf{Benchmark} 
  & \textbf{Resp Len.} 
  & \textbf{Tokens} 
  & \textbf{Cost (\$)} 
  & \textbf{\#Sents} 
  & \textbf{\#Claims} 
  & \textbf{Supp.} 
  & \textbf{Pre\_verif.\%} 
  & \textbf{P} 
  & \textbf{F1@K'} \\
\midrule
\endhead
\bottomrule
\endfoot
  \multirow{6}{*}{\makecell{gemini{-}2.0{-}flash\\{-}thinking{-}exp}} & factscore\_bio & 576.2 & 5266.0 & 0.035 & 27.0 & 13.0 & 11.2 & 0.646 & 0.845 & 0.783 \\
   & longfact & 1981.0 & 5972.0 & 0.025 & 101.4 & 32.0 & 30.2 & 0.912 & 0.95 & 0.851 \\
   & factcheck-bench & 518.6 & 11832.0 & 0.083 & 28.0 & 21.6 & 17.8 & 0.463 & 0.824 & 0.711 \\
   & expertqa & 376.0 & 3738.0 & 0.021 & 19.0 & 17.0 & 16.333 & 0.725 & 0.955 & 0.879 \\
   & hellobench & 926.0 & 6921.0 & 0.047 & 42.6 & 13.8 & 11.8 & 0.551 & 0.82 & 0.789 \\
\cmidrule(lr){2-11}
   & \textbf{Avg} & 875.56 & 6745.8 & 0.042 & 43.6 & 19.48 & 17.467 & 0.659 & 0.879 & \textbf{0.803} \\
\midrule
  \multirow{6}{*}{gpt{-}4o} & factscore\_bio & 205.2 & 17529.0 & 0.126 & 9.35 & 8.25 & 6.85 & 0.527 & 0.784 & 0.741 \\
   & longfact & 565.75 & 19235.0 & 0.096 & 31.7 & 18.5 & 18.25 & 0.846 & 0.986 & 0.847 \\
   & factcheck-bench & 225.0 & 13392.0 & 0.078 & 12.95 & 5.25 & 4.5 & 0.552 & 0.831 & 0.796 \\
   & expertqa & 461.8 & 26822.0 & 0.173 & 26.85 & 12.15 & 11.8 & 0.547 & 0.971 & 0.839 \\
   & hellobench & 734.6 & 20877.0 & 0.115 & 47.4 & 18.0 & 17.5 & 0.806 & 0.966 & 0.708 \\
\cmidrule(lr){2-11}
   & \textbf{Avg} & 438.47 & 19571.0 & 0.118 & 25.65 & 12.43 & 11.78 & 0.656 & 0.908 & \textbf{0.786} \\
\midrule
  \multirow{6}{*}{\makecell{gpt{-}3.5\\{-}turbo{-}0125}} & factscore\_bio & 182.2 & 6085.0 & 0.048 & 8.4 & 9.8 & 5.6 & 0.327 & 0.6 & 0.487 \\
   & longfact & 367.6 & 3406.0 & 0.015 & 16.4 & 13.6 & 13.6 & 0.868 & 1.0 & 0.809 \\
   & factcheck-bench & 134.8 & 2027.0 & 0.011 & 8.0 & 5.4 & 5.0 & 0.778 & 0.958 & 0.851 \\
   & expertqa & 394.8 & 6213.0 & 0.042 & 15.2 & 10.4 & 10.0 & 0.462 & 0.94 & 0.874 \\
   & hellobench & 425.0 & 5615.0 & 0.039 & 18.8 & 9.4 & 8.6 & 0.447 & 0.943 & 0.889 \\
\cmidrule(lr){2-11}
   & \textbf{Avg} & 300.88 & 4669.2 & 0.031 & 13.36 & 9.72 & 8.56 & 0.576 & 0.888 & \textbf{0.782} \\
\midrule
  \multirow{6}{*}{gemini{-}2.0{-}flash} & factscore\_bio & 411.6 & 5768.0 & 0.035 & 21.4 & 21.4 & 18.4 & 0.776 & 0.878 & 0.892 \\
   & longfact & 1461.2 & 5927.0 & 0.03 & 80.2 & 25.2 & 24.0 & 0.865 & 0.917 & 0.75 \\
   & factcheck-bench & 440.8 & 3671.0 & 0.021 & 23.0 & 10.6 & 9.4 & 0.774 & 0.865 & 0.815 \\
   & expertqa & 582.0 & 4761.0 & 0.023 & 35.6 & 19.2 & 18.6 & 0.854 & 0.963 & 0.66 \\
   & hellobench & 887.0 & 11291.0 & 0.067 & 42.2 & 24.6 & 22.8 & 0.642 & 0.917 & 0.789 \\
\cmidrule(lr){2-11}
   & \textbf{Avg} & 756.52 & 6283.6 & 0.035 & 40.48 & 20.2 & 18.64 & 0.782 & 0.908 & \textbf{0.781} \\
\midrule
  \multirow{6}{*}{DeepSeek{-}V3} & factscore\_bio & 204.15 & 15261.0 & 0.096 & 11.55 & 9.95 & 8.5 & 0.709 & 0.791 & 0.764 \\
   & longfact & 582.45 & 32351.0 & 0.161 & 39.6 & 27.55 & 25.3 & 0.817 & 0.911 & 0.784 \\
   & factcheck-bench & 295.05 & 22235.0 & 0.137 & 19.1 & 13.75 & 12.5 & 0.695 & 0.885 & 0.802 \\
   & expertqa & 534.5 & 34904.0 & 0.215 & 39.55 & 17.05 & 15.9 & 0.601 & 0.94 & 0.818 \\
   & hellobench & 665.1 & 27567.0 & 0.156 & 51.2 & 20.1 & 19.05 & 0.761 & 0.876 & 0.698 \\
\cmidrule(lr){2-11}
   & \textbf{Avg} & 456.25 & 26463.6 & 0.153 & 32.2 & 17.68 & 16.25 & 0.717 & 0.881 & \textbf{0.773} \\
\midrule
  \multirow{7}{*}{\makecell{Qwen2.5{-}7B\\{-}Instruct}} & factscore\_bio & 235.0 & 10485.0 & 0.072 & 11.769 & 7.846 & 5.462 & 0.569 & 0.723 & 0.723 \\
   & factcheck-bench & 302.75 & 15589.0 & 0.086 & 17.95 & 8.2 & 7.2 & 0.695 & 0.844 & 0.759 \\
   & expertqa & 430.2 & 22185.0 & 0.136 & 37.0 & 11.85 & 11.25 & 0.629 & 0.939 & 0.814 \\
   & HelloBench & 811.5 & 19431.0 & 0.092 & 53.1 & 19.45 & 18.8 & 0.866 & 0.943 & 0.72 \\
   & longfact\_concepts & 675.8 & 10377.0 & 0.048 & 37.8 & 20.9 & 20.2 & 0.866 & 0.955 & 0.759 \\
   & longfact\_objects & 522.3 & 11405.0 & 0.061 & 28.6 & 19.2 & 17.1 & 0.807 & 0.882 & 0.819 \\
\cmidrule(lr){2-11}
   & \textbf{Avg} & 496.258 & 14912.0 & 0.082 & 31.037 & 14.574 & 13.335 & 0.739 & 0.881 & \textbf{0.766} \\
\midrule
  \multirow{6}{*}{DeepSeek{-}R1} & factscore\_bio & 297.15 & 22905.0 & 0.171 & 14.45 & 13.0 & 8.9 & 0.585 & 0.693 & 0.655 \\
   & longfact & 540.25 & 36893.0 & 0.202 & 36.1 & 28.8 & 26.3 & 0.788 & 0.873 & 0.659 \\
   & factcheck-bench & 303.15 & 17376.0 & 0.102 & 20.8 & 11.65 & 10.3 & 0.73 & 0.861 & 0.804 \\
   & expertqa & 624.05 & 44394.0 & 0.275 & 36.15 & 20.8 & 18.05 & 0.575 & 0.851 & 0.757 \\
   & hellobench & 675.85 & 28330.0 & 0.162 & 54.15 & 20.8 & 19.25 & 0.76 & 0.897 & 0.661 \\
\cmidrule(lr){2-11}
   & \textbf{Avg} & 488.09 & 29979.6 & 0.182 & 32.33 & 19.01 & 16.56 & 0.688 & 0.835 & \textbf{0.707} \\
\midrule
  \multirow{6}{*}{gpt{-}4{-}turbo} & factscore\_bio & 155.6 & 1389.0 & 0.002 & 7.0 & 5.8 & 2.8 & 0.483 & 0.587 & 0.635 \\
   & longfact & 468.0 & 2193.0 & 0.003 & 26.0 & 19.0 & 16.8 & 0.884 & 0.863 & 0.772 \\
   & factcheck-bench & 149.6 & 937.0 & 0.002 & 8.0 & 6.2 & 3.6 & 0.581 & 0.655 & 0.709 \\
   & expertqa & 402.2 & 1993.0 & 0.003 & 34.4 & 14.4 & 10.4 & 0.722 & 0.698 & 0.698 \\
   & hellobench & 576.4 & 1837.0 & 0.003 & 26.8 & 12.8 & 8.0 & 0.625 & 0.567 & 0.633 \\
\cmidrule(lr){2-11}
   & \textbf{Avg} & 350.36 & 1669.8 & 0.003 & 20.44 & 11.64 & 8.32 & 0.659 & 0.674 & \textbf{0.689} \\
\midrule
  \multirow{6}{*}{o1} & factscore\_bio & 402.8 & 2324.0 & 0.003 & 19.0 & 17.0 & 10.8 & 0.635 & 0.591 & 0.58 \\
   & longfact & 1450.8 & 3799.0 & 0.005 & 62.6 & 29.2 & 24.2 & 0.829 & 0.798 & 0.678 \\
   & factcheck-bench & 307.4 & 1130.0 & 0.002 & 17.2 & 7.0 & 5.2 & 0.743 & 0.607 & 0.634 \\
   & expertqa & 336.6 & 1757.0 & 0.003 & 14.0 & 10.6 & 9.2 & 0.868 & 0.872 & 0.78 \\
   & hellobench & 1004.4 & 3150.0 & 0.004 & 46.8 & 25.8 & 21.6 & 0.845 & 0.796 & 0.632 \\
\cmidrule(lr){2-11}
   & \textbf{Avg} & 700.4 & 2432.0 & 0.003 & 31.92 & 17.92 & 14.2 & 0.784 & 0.733 & \textbf{0.661} \\
\midrule
  \multirow{5}{*}{\makecell{Qwen{-}7B\\{-}Chat{-}new}} & longfact & 389.4 & 2074.0 & 0.003 & 19.0 & 15.6 & 12.4 & 0.795 & 0.793 & 0.671 \\
   & factcheck-bench & 146.8 & 1103.0 & 0.002 & 7.4 & 9.6 & 3.2 & 0.333 & 0.407 & 0.421 \\
   & expertqa & 375.8 & 2093.0 & 0.003 & 23.0 & 18.4 & 15.8 & 0.859 & 0.844 & 0.79 \\
   & hellobench & 483.0 & 1604.0 & 0.003 & 21.8 & 12.2 & 9.0 & 0.738 & 0.674 & 0.682 \\
\cmidrule(lr){2-11}
   & \textbf{Avg} & 348.75 & 1718.5 & 0.003 & 17.8 & 13.95 & 10.1 & 0.681 & 0.679 & \textbf{0.641} \\
\midrule
  \multirow{6}{*}{o1{-}mini} & factscore\_bio & 383.2 & 1424.0 & 0.003 & 17.0 & 5.8 & 3.0 & 0.517 & 0.483 & 0.562 \\
   & longfact & 1769.0 & 4338.0 & 0.006 & 97.0 & 35.0 & 33.6 & 0.96 & 0.938 & 0.689 \\
   & factcheck-bench & 427.8 & 2149.0 & 0.003 & 27.4 & 16.8 & 12.0 & 0.714 & 0.738 & 0.64 \\
   & expertqa & 494.4 & 2420.0 & 0.004 & 41.4 & 23.4 & 20.2 & 0.863 & 0.799 & 0.662 \\
   & hellobench & 1380.4 & 3410.0 & 0.005 & 75.4 & 29.8 & 22.4 & 0.752 & 0.66 & 0.548 \\
\cmidrule(lr){2-11}
   & \textbf{Avg} & 890.96 & 2748.2 & 0.004 & 51.64 & 22.16 & 18.24 & 0.761 & 0.724 & \textbf{0.620} \\
\midrule
  \multirow{6}{*}{qwen2.5{-}72b{-}instruct} & factscore\_bio & 370.6 & 2298.0 & 0.003 & 17.0 & 16.6 & 11.8 & 0.711 & 0.669 & 0.609 \\
   & longfact & 782.8 & 3130.0 & 0.004 & 54.0 & 24.6 & 23.6 & 0.959 & 0.933 & 0.752 \\
   & factcheck-bench & 334.2 & 1744.0 & 0.003 & 18.8 & 10.8 & 5.4 & 0.5 & 0.474 & 0.439 \\
   & expertqa & 518.2 & 2272.0 & 0.003 & 50.6 & 19.4 & 13.8 & 0.711 & 0.668 & 0.659 \\
   & hellobench & 872.6 & 2569.0 & 0.004 & 42.2 & 21.4 & 15.2 & 0.71 & 0.655 & 0.6 \\
\cmidrule(lr){2-11}
   & \textbf{Avg} & 575.68 & 2402.6 & 0.003 & 36.52 & 18.56 & 13.96 & 0.718 & 0.68 & \textbf{0.612} \\
\midrule
  \multirow{6}{*}{gpt{-}4o{-}mini} & factscore\_bio & 201.2 & 1634.0 & 0.002 & 8.6 & 7.6 & 3.0 & 0.395 & 0.341 & 0.432 \\
   & longfact & 751.8 & 3077.0 & 0.004 & 42.4 & 26.0 & 22.4 & 0.862 & 0.859 & 0.789 \\
   & factcheck-bench & 252.6 & 960.0 & 0.002 & 13.6 & 4.4 & 1.6 & 0.364 & 0.298 & 0.307 \\
   & expertqa & 486.6 & 2237.0 & 0.003 & 35.4 & 17.4 & 12.4 & 0.713 & 0.692 & 0.509 \\
   & hellobench & 829.2 & 2491.0 & 0.004 & 39.0 & 18.6 & 12.4 & 0.667 & 0.565 & 0.558 \\
\cmidrule(lr){2-11}
   & \textbf{Avg} & 504.28 & 2079.8 & 0.003 & 27.8 & 14.8 & 10.36 & 0.6 & 0.551 & \textbf{0.519} \\
\bottomrule
\end{longtable}
\end{center}

\section{\name\ Annotation Protocol}
\label{appdx:anntation-protocol}


# Annotation Protocol for FastFact

## Background

In this annotation task, you will review and revise a factuality evaluation pipeline of long-form generations from 4 sampled models on a few questions. To evaluate a model generation, the evaluation pipeline consists of three subprocesses: (1) claim extraction and preverification, (2) evidence collection, and (3) claim verification.

(1) **claim extraction and preverification**: A claim extractor decomposes a long-form generation into `atomic claims', by extracting verifiable factual statements from the model generation. If there is no extractable claim in the response, the extraction output would be "No verifiable claim.";
(2) **evidence collection**: retrieve or search for the evidence related to each claim;
(3) **claim verification**: verify each claim's factual correctness based on the evidence.

Lastly, each claim's classification label is aggregated together as a factuality score.

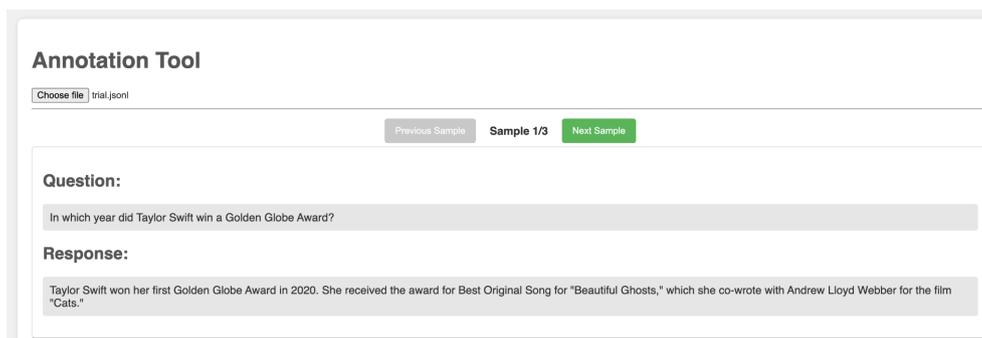

## Tasks

For each sample, you will review the outputs of (1) claim extraction and preverification and (3) claim verification subprocesses of the evaluation pipeline, and perform two tasks across two screens:

TASK 1: REVISE EXTRACTED CLAIMS

> During this subprocess, the extractor extracts <u>as many verifiable atomic factual claims</u> mentioned in the model generation as possible. **Verifiable claims** describe a single event or state with all necessary modifiers (e.g., spatial, temporal, or relative clauses) that help denote entities or contextualize events in the real world, and can be verifiable given reliable external world knowledge (e.g., via Wikipedia).

- **Your input**: question + model response + a list of extracted claims.

- **Your task**: Based on the model generation and existing extracted claims, Task 1 is to annotate the quantity and quality of the claim extraction subprocess, by removing/adding/revising claims.

(1) **Remove/Add.** For a question-response pair, there should be a certain number of atomic facts that are included in the response, not more or less. Therefore, you should remove those that are {redundant claims, unextracted claims, not a (verifiable) claim}, and add missing claims that were contained in the model response.

    (a) **Remove**: the two following types of claims should be removed. If removed a claim, you will be asked right away to label the reason for removal:
- Redundant: The meaning of the current claim overlaps exactly with part of another claim.
- Not a claim: The claim is about a story, personal experiences, hypotheticals (e.g., "would be" or subjunctive), subjective statements (e.g., opinions), future predictions, suggestions, advice, instructions, and other such content.

**Original Claim:** *"Beautiful Ghosts" was written for the film "Cats."*

Reason for removal: [-- Select reason -- ▾]

[Confirm Removal Label] [Cancel Removal]

    (b) **Add**: Add any missing verifiable claims in the exact sequence order corresponding to the response text.

**Original Claim:** *Taylor Swift won her first Golden Globe Award in 2020.*

```
Taylor Swift won her first Golden Globe Award in 2020.
```

[Remove Claim]

[+ Add New Claim]

**New Claim (added by annotator):**

```
Enter claim text here...
```

[Cancel Add Claim]

[+ Add New Claim]

(2) **Revise.** A verifiable claim should be context-independent and easily grounded with evidence search. Therefore, please make revisions to the extracted claim if a claim statement:

    (a) Is ambiguous or contains equivocal information. Each extracted claim should also be describing either one single event (e.g., "Nvidia is founded in 1993 in Sunnyvale, California, U.S.") or single state (e.g., "UMass Amherst has existed for 161 years.") with necessary time and location information.

    (b) Is contextually dependent. Revise the contextually dependent parts (e.g., pronouns) into verifiable named entities, based on the context information in the model response and the question. Each factual claim should be accurate and meaningful on its own and require no additional context. This means that each claim must be situated within relevant temporal information and location whenever available, and all entities in the claim must

be referred to by name but not pronoun. Use the name of entities (e.g., 'Edward Jenner') rather than definite noun phrases (e.g., 'the doctor') whenever possible. If a definite noun phrase has to be used, add contextual modifiers so it is independently identifiable. Keep each factual claim to one sentence with zero or at most one embedded clause.

Note that the meaning of the extracted claims should adhere to the original model response, regardless of claim correctness. Your revision should not alter the meaning of the original response.

**Original Claim:** *Taylor Swift received the Golden Globe Award for Best Original Song for "Beautiful Ghosts" in 2020.*

> Taylor Swift received the Golden Globe Award for Best Original Song for "Beautiful Ghosts" in 2020.

[Remove Claim]

[+ Add New Claim]

TASK 2. VERIFY THE REVISED CLAIMS

- **Your input**: Each (revised) claim + claim label + (OPTIONAL) searched evidence
- **Your task**: Decide whether you agree with the verification label for each revised claim; if not, give the correct verification label.
- **Your annotation labels:**
    - Agree
    - Disagree: if so, give the correct verification label based on your judgement.
        i. Irrelevant: A claim is completely irrelevant to the provided question.
        ii. Supported: A claim is supported if everything in the claim is supported and nothing is contradicted by the provided (or your own) search results. Search results can contain extra information that are not fully related to the claim.
        iii. Refuted: A claim is contradicted if a part of the claim is contradicted by a part of the provided (or your own) search results, and if no search result that supports the same part.
        iv. Conflicting evidence: A claim is controversial if a part of a claim cannot be verified by the provided (or your own) search results, because the claim is supported and contradicted by different mixed pieces of evidence in the search results.
        v. Not enough evidence: A claim is inconclusive if a part of a claim cannot be verified by the provided (or your own) search results, because there is not sufficient information in the search results related to the claim to make a verification.

```
Is the Met Gala Couples Curse real? What to know about the Met Gala
Benny Blanco and Selena Gomez
Ben Affleck's new aura
John Mulaney is blowing up late night
Liam Hemsworth transformation
```

**Reasoning for System's Label:**

The claim states that Taylor Swift won her first Golden Globe Award in 2020. The evidence provided won. Specifically, Evidence 3 and Evidence 8 clearly state that Taylor Swift has never won a Gold nominated for "Beautiful Ghosts" at the 2020 ceremony, but there is no mention of a win. Thus, th evidence, which shows that she has not won any Golden Globe awards as of the latest updates.

**System's Verification Label:** refuted

Verification Agreement: ○ Agree  ● Disagree

Correct Label  [✓ Select correct label ▾]
    supported
    refuted
**Original Clai**   conflicting evidence   the Golden Globe Award for Best Original Song for "Beautiful Ghosts"
                    not enough evidence
**Revised Clai**    irrelevant             d the Golden Globe Award for Best Original Song for "Beautiful Ghosts"

- **Decision framework per claim**:
    a. Verifying a claim without searched evidence provided / Verifying your newly added claim:
        - You will decide whether to agree with the verification label without being provided any evidence. Where there is no explicit search evidence, the displayed verification labels would be only be [Irrelevant (i), Supported (ii), Unsupported (iii, iv, v)], with the Unsupported label encompassing the last three situations (iii, iv, v).
        - Manually search webs/wikipedia for as much evidence as needed, to decide the correct label of the claim. If you disagree, please choose from the 5 annotation labels above. You can click the 🔍 button beside the (revised) claim to automatically search about the claim on Google.

    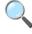

    b. Verifying a claim with searched evidence provided:
        - You will judge the model's verification label and its reasoning based on the searched evidence. The displayed verification labels would be from the same list of annotation labels above (i, ii, iii, iv, v).
        - Consider ONLY the evidence presented to decide the correct label of the current claim. Ignore any external knowledge that contradicts the evidence. If you disagree, please choose the correct label from the 5 annotation labels above.



\section{\name\ Fewshot Prompt for Claim Extraction \& Pre-Verification}
\label{appdx:prompt-extraction}
Below is the few-shot prompt template for step 1: claim extraction and pre-verification. The prompt explicitly declares two subtasks: (a) decomposing or extracting a generation into verifiable atomic claims and (b) pre-verifying the factual correctness of each claim. The instruction for subtask (a) at the beginning is adapted from \citet{veriscore}.

\begin{tcolorbox}[breakable]
\obeylines

\small
You are trying to verify how factual a response to a question or request is. To do so, you need to perform the following two steps.\\

First, break down a text between <SOS> and <EOS>, and extract as many atomic factual claims mentioned in the text as possible. An atomic factual claim should be verifiable against reliable external world knowledge (e.g., via Wikipedia). Any story, personal experiences, hypotheticals (e.g., "would be" or subjunctive), subjective statements (e.g., opinions), suggestions, advice, instructions, and other such content should not be extracted. Note that biographical, historical, scientific, and other such information are not personal experiences or stories, so they should be extracted. Each extracted claim should be describing a single event (e.g., "Nvidia is founded in 1993 in Sunnyvale, California, U.S."), single state (e.g., "UMass Amherst has existed for 161 years."), or a concrete piece of information (e.g., "A typical human diploid cell contains 46 chromosomes (23 pairs)."). Quotations should be extracted verbatim with the source when available.\\
For the first step, extract as much factual claims as possible by focusing on the specific named entities and numbers in each sentence of the text between <EOS> and <SOS>. Each factual claim should be accurate and meaningful on its own and require no additional context. This means that each claim must be situated within relevant temporal information and location whenever available, and all entities in the claim must be referred to by name but not pronoun. Use the name of entities (e.g., 'Edward Jenner') rather than definite noun phrases (e.g., 'the doctor') whenever possible. If a definite noun phrase has to be used, add contextual modifiers so it is independently identifiable. Keep each claim as one sentence with at most one embedded clause.
You do not need to justify what you extract. Simply extract the atomic factual claims following the requirements, regardless of claim correctness. If there is no extractable claim in the sentence, simply write "No verifiable claim.".\\
Second, act as an evaluator and verify each extracted claim. You should first check whether each extracted claim is relevant in answering the provided question. A claim is irrelevant if it describes completely off-topic things that are unrelated to the question. If the claim is irrelevant, simply label it as IRRELEVANT; If the claim is relevant, continue to verify the factual correctness of the claim with respect to the real world based on your own knowledge.\\

For the second step, choose one out of the following labels for each factual claim: 
- If a claim is completely irrelevant to the provided question, label it as IRRELEVANT;
- If a claim is relevant to the question and you are fully confident it is factually incorrect (e.g., "The Earth orbits around the Moon."), label it as NON-SUPPORTED; 
- If a claim is relevant to the question and you are fully confident it is true and factually accurate (e.g., "The Earth orbits around the Sun."), label it as SUPPORTED; 
- If a claim is relevant to the question and it is likely true, label it as LIKELY SUPPORTED;
- If a claim is relevant to the question and it is likely false, label it as LIKELY NON-SUPPORTED;
- If a claim is relevant to the question, but you are unsure of its factual correctness, or if the claim is equivocal or controversial and requires further evidence, label it as UNSURE.
Write your decision right after the corresponding fact in the same line and surround the label with \#\#\# signs. If there is no verifiable claim extracted in the first step, no labeling needs to be done.\\

Here are some examples: \\
Question: In which year did Taylor Swift win a Golden Globe Award?
Response: <SOS>Taylor Swift won her first Golden Globe Award in 2020. She received the award for Best Original Song for \"Beautiful Ghosts\", which she co-wrote with Andrew Lloyd Webber for the film \"Cats.\"<EOS>
Claims:
- Taylor Swift won her first Golden Globe Award in 2020. \#\#\#NON-SUPPORTED\#\#\#
- Taylor Swift received the Golden Globe Award for Best Original Song for "Beautiful Ghosts" in 2020. \#\#\#NON-SUPPORTED\#\#\#
- Taylor Swift co-wrote "Beautiful Ghosts" with Andrew Lloyd Webber. \#\#\#LIKELY SUPPORTED\#\#\#
- "Beautiful Ghosts" was written for the film "Cats." \#\#\#SUPPORTED\#\#\# \\

Question: What NASA programs would support our college in starting a robotics program?
Response: <SOS>Here are a few:
1. NASA Robotics Alliance Project (RAP): This program provides educational resources and support for robotics teams, including college-level teams, that are participating in NASA robotics competitions.<EOS>
2. NASA Minority University Research and Education Project (MUREP): This program provides funding and resources for colleges and universities with a significant minority student population to develop research and education programs in STEM fields, including robotics.
Claims:
- NASA has a program called NASA Robotics Alliance Project (RAP). \#\#\#SUPPORTED\#\#\#
- NASA Robotics Alliance Project provides educational resources and supports for robotics teams. \#\#\#SUPPORTED\#\#\#
- NASA Robotics Alliance Project provides supports educational resources and for college-level teams that are participating in NASA robotics competitions. \#\#\#UNSURE\#\#\# \\

Question: How can I find a law firm that specializes in copyright related matters?
Response: <SOS>There are a few ways:
1. Online search: Search online using keywords like \"copyright law firm\" or \"copyright lawyer\" along with your location.
2. Ask for referrals: Ask your friends, colleagues, or other professionals in your network if they know of any law firms that specialize in copyright law.<EOS>
Claims:
No verifiable claim. \\

Question: I want Ubutu but i have a question first. I'm not new to Ubuntu, my friends use it. i never had a chance to use it on my own PC i'm running on a window 8 and it has no info like product keys and such it was on the box and this is a hand me down PC. My question is do i need those for the installation, if so how do i retrieve this info.
Response: <SOS>You might need to make a windows recovery disk. You need a windows recovery disk in the event you have a problem with windows.<EOS>
Claims:
- One needs a windows recovery disk if one has a problem with windows when installing Ubuntu. \#\#\#LIKELY SUPPORTED\#\#\#\\

Question: What happens to you if you eat watermelon seeds?
Response: <SOS>If you accidentally or intentionally swallow a few watermelon seeds while eating the fruit, there is no need to worry. You can safely consume watermelon seeds as they even possess some beneficial properties such as being a good source of protein.<EOS>
Claims:
- Consuming a few watermelon seeds while eating the fruit is not dangerous for the human body. \#\#\#SUPPORTED\#\#\#
- Watermelon seeds are a good source of protein. \#\#\#LIKELY NON-SUPPORTED\#\#\#\\

Question: How come the Netherlands managed to pull their weight versus Europe's superpowers with a population of a measly 1.5 million?
The trading empire of the Dutch Republic would be impressive alone for it's achievements in trade and warfare, but it's astounding when you realize that by 1600, the Netherlands had 1.5m for population! Compare that to 18m of France, or 9m approx of Spain, or 5.6m by England.
I get that they were very sophisticated and had advanced commerce, production and politics. But still... 1.5 million? How is that enough manpower to arm enough soldiers on land to prevent invasions, and sailors to operate a huge fleet? And given the immense wealth and manpower of Spain and Portugal compared to that of the Dutch - what prevented them from just dropping in on Amsterdam and burn it so to keep them out of their business? Or maybe drop in on Calais and march on land if they didn't have enough naval power.
Response: <SOS>I'm writing a paper on the Dutch army around that time at this very moment.
I'll try to answer a couple of your questions. One thing to know is that the military strength of the Netherlands varied greatly in short periods of time. I'll mostly be focussing on the period around 1600, because that's the year you mentioned in your question.<EOS>
Another thing to know is that the 'manpower' of the Netherlands itself doesn't really translate well into actual figures for the Dutch army.
Claims:
- The military strength of the Netherlands varied greatly in short periods of time around 1600. \#\#\#LIKELY SUPPORTED\#\#\#\\

Question: What is the smallest positive integer that cannot be expressed as the sum of 7s, 11s, and 13s?
Response: <SOS>This is a classic problem in number theory known as the Frobenius coin problem or the coin problem of Frobenius. The formula for the smallest integer that cannot be expressed as the sum of two relatively prime integers a and b is ab - a - b. In this case, the integers 7, 11, and 13 are pairwise relatively prime, so we can apply this formula to find the smallest integer that cannot be expressed as a sum of 7s, 11s, and 13s.

Using the formula, we get: 7 x 11 - 7 - 11 = 59 
So the smallest positive integer that cannot be expressed as the sum of 7s, 11s, and 13s is 59.<EOS>
Claims:
- The Frobenius coin problem formula for the smallest integer that cannot be expressed as the sum of two relatively prime integers a and b is ab - a - b. \#\#\#NON-SUPPORTED\#\#\#\\

Extract *verifiable atomic* factual claims and then judge them based on your knowledge.\\

\{Response Chunk\}
Claims:

\end{tcolorbox}

\section{\name\ Fewshot Prompt for Claim Verification}
\label{appdx:prompt-verification}
\begin{tcolorbox}[breakable]
\obeylines

\small
You need to judge whether a claim is supported or refuted by the provided evidence extracted from searched web pages. Based on the claim and the searched evidence, first reason about which verification label to assign the claim, then write your final decision. Surround the label with \#\#\# signs.\\

Below are the definitions of the five categories:\\

\#\#\#supported\#\#\#: Everything in the claim is supported and nothing is refuted by the search results. Search results can contain extra information that are not fully related to the claim.
\#\#\#refuted\#\#\#: A part of the claim is refuted by the evidence in search results, and there is no evidence that supports the same part.
\#\#\#conflicting evidence\#\#\#: A part of a claim cannot be verified by the search results, because the claim is supported and refuted by different mixed pieces of evidence in the search results.
\#\#\#not enough evidence\#\#\#: A part of a claim cannot be verified by the search results, because there is not sufficient information in the search evidence related to the claim to make a verification.
\#\#\#unverifiable\#\#\#: A part of a claim cannot be verified, because the claim is not a claim that states a specific fact or event. For example, a claim is unverifiable if the claim itself is ambiguous, a question, or an opinion.\\

Here are some examples. Follow the format in the examples:\\

Claim: Lenny and Carl are depicted as a gay couple on The Simpsons.
Searched Results: Evidence 1
Source Title: Are Simpsons' Carl \& Lenny Gay? Every Clue To Their Relationship
Content: the Springfield Nuclear Power Plant, and along with Barney Gumble, they are his best friends. Lenny and Carl are inseparable, and their relationship has sparked many questions, as they are often portrayed as a couple, but there have been other details hinting at the contrary.  Remove Ads The Simpsons: Lenny \& Carl’s Relationship Explained Leonard “Lenny” Leonard made his first appearance in season 1’s episode “Life on the Fast Lane”, and he’s a Technical Supervisor at the Springfield Nuclear Power Plant. He has a master’s degree in nuclear physics but he’s portrayed as a simple and often naive man. Carlton “Carl” Carlson made his debut in the following episode, “Homer’s Night Out”, and he’s Safety Operations Supervisor at the Power Plant. Very much like Lenny, he has a master’s degree in nuclear physics, and it has been implied that he was a war hero. Lenny and Carl are inseparable, and the series has hinted at an actual romantic relationship between them multiple times. Remove Ads

Lenny and Carl being a couple is a running gag in The Simpsons, and the writers often play with it with double entendre or through visual jokes.

Evidence 2
Source Title: Are Simpsons' Carl \& Lenny Gay? Every Clue To Their Relationship
Content: within their hearts and souls (and Carl saw his own face in the stars). Lenny also published a newspaper called The Lenny-Saver with the headline “The Truth About Carl: He’s Great”, of which he was very proud, and a framed photograph of Carl can be seen at his home. Other occasions that have pointed at a romantic relationship between Lenny and Carl are when Marge’s popsicle sticks sculptures were destroyed and Lenny and Carl’s were mashed together, with Lenny saying that he didn’t know “where Carls ends and I begin!” and Carl quickly replied “it’s statements like that that make people think we’re gay”. One time at the Springfield Baseball Stadium, while watching the Kiss Cam, Lenny reminded Carl when they used to do that... with their respective girlfriends. Though they have been shown holding hands, there have also been references that point at them being straight, such as both having mistresses. Lenny and Carl’s relationship in The Simpsons will be whatever each viewer wants it to be, as the series will surely keep playing with it and won’t confirm if they are a couple or just friends.

Evidence 3
Source Title: Are Simpsons' Carl \& Lenny Gay? Every Clue To Their Relationship
Content: relationship between Lenny and Carl are when Marge’s popsicle sticks sculptures were destroyed and Lenny and Carl’s were mashed together, with Lenny saying that he didn’t know “where Carls ends and I begin!” and Carl quickly replied “it’s statements like that that make people think we’re gay”. One time at the Springfield Baseball Stadium, while watching the Kiss Cam, Lenny reminded Carl when they used to do that... with their respective girlfriends. Though they have been shown holding hands, there have also been references that point at them being straight, such as both having mistresses. Lenny and Carl’s relationship in The Simpsons will be whatever each viewer wants it to be, as the series will surely keep playing with it and won’t confirm if they are a couple or just friends.

Evidence 4
Source Title: r/FanTheories on Reddit: [The Simpsons] Lenny and Carl aren't ambiguously gay (originally)
Content: Skip to main content        [The Simpsons] Lenny and Carl aren't ambiguously gay (originally) : r/FanTheories
Background: "The Simpsons" famously started as a parody and deconstruction of existing tropes (especially family sitcoms): Bart is Dennis if he was actually a Menace, Moe's Tavern is a grittier Cheers, Homer is a classic sitcom dad robbed of all intellectual and moral authority. After the first 8-10 seasons of the show, the "wholesome" sitcoms it was parodying were gradually forgotten and the best showrunners and writers moved on, and many of the characters moved from 3d characters and meta-commentary to 2d stereotypes (a process often called "Flanderization"). Theory: Lenny and Carl started out as a parody of background characters who always appear as a pair: Crabbe and Goyle or Fred and George in "Harry Potter", Kuby and Huell in "Breaking Bad", most of the dwarves with rhyming names in "The Hobbit", etc.

Evidence 5
Source Title: Are Simpsons' Carl \& Lenny Gay? Every Clue To Their Relationship
Content: Such is the case of two of Homer’s closest friends: Lenny and Carl. Related: Why Simpsons Characters Are Blonde In The First Season Homer met Lenny and Carl way before they began working at the Springfield Nuclear Power Plant, and along with Barney Gumble, they are his best friends. Lenny and Carl are inseparable, and their relationship has sparked many questions, as they are often portrayed as a couple, but there have been other details hinting at the contrary. Remove Ads

Reasoning: The claim states that Lenny and Carl are depicted as a gay couple on The Simpsons. The evidence highlights that their relationship is intentionally ambiguous and played for humor. Multiple sources (Evidence 1, 2, 3, 5) note that the show hints at a romantic relationship through jokes, double entendres, and visual gags (e.g., holding hands, mashed sculptures). However, other evidence (Evidence 2, 3) explicitly refutes this, such as Carl stating, "it’s statements like that that make people think we’re gay," and references to them having girlfriends or mistresses. Additionally, Evidence 4 suggests their pairing is a parody of background character tropes, not necessarily romantic. The search results include both supporting and refuting claims about their relationship, and the show deliberately avoids confirmation. Thus, the claim is neither fully supported nor fully refuted but involves mixed evidence.
Decision: \#\#\#conflicting evidence\#\#\#\\

Claim: The 2026 FIFA World Cup final will be held at MetLife Stadium in New Jersey, USA.
Searched Results: Evidence 1
Source Title: 2026 World Cup final set for MetLife Stadium, USA kicks off play in L.A. - ESPN
Content: The 2026 World Cup final will be held at MetLife Stadium in East Rutherford, New Jersey, on July 19, world soccer governing body FIFA announced on Sunday for the tournament being hosted by the United States, Mexico and Canada.

Evidence 2
Source Title: The 2026 World Cup final will take place at New Jersey's MetLife Stadium
Content: Accessibility links

Skip to main content
Keyboard shortcuts for audio player

2026 World Cup final to take place at New Jersey MetLife Stadium The complex home to the New York Jets and Giants and located in East Rutherford, N.J., will be renamed the New York New Jersey Stadium during the July 19 final. Sports

The 2026 World Cup final will take place at New Jersey's MetLife Stadium

February 4, 20249:05 PM ET

The 2026 World Cup final will be played at MetLife Stadium in East Rutherford, N.J., on July 19, FIFA announced on Sunday. Seth Wenig/AP hide caption toggle caption Seth Wenig/AP

The 2026 World Cup final will be played at MetLife Stadium in East Rutherford, N.J., on July 19, FIFA announced on Sunday. Seth Wenig/AP
The 2026 World Cup final will take place at New Jersey's MetLife Stadium to cap a tournament set in cities across the U.S., Canada and Mexico, soccer's international governing body FIFA announced Sunday. The final will be played on July 19 at the East Rutherford, N.J., stadium. Mexico City will host the opener of the 104-game tournament at Estadio Azteca on June 11.

Evidence 3
Source Title: 2026 World Cup final will be played at MetLife Stadium in New Jersey
Content: The 2026 World Cup final will be played at MetLife Stadium in East Rutherford, New Jersey, on July 19. FIFA made the announcement Sunday, allocating the opener of the 39-day tournament to Mexico City’s Estadio Azteca on June 11 and the finale to the home of the NFL’s New York Jets and Giants.

Evidence 4
Source Title: 2026 FIFA World Cup final - Wikipedia
Content: Jump to content Article Talk English Read Edit View history Tools Tools move to sidebar hide Actions Read Edit View history General What links here Related changes Upload file Special pages Permanent link Page information Cite this page Get shortened URL Download QR code Edit interlanguage links Expand all Print/export Download as PDF Printable version In other projects Wikidata item From Wikipedia, the free encyclopedia Soccer match in East Rutherford, New Jersey Football match 2026 FIFA World Cup final Aerial view of MetLife Stadium in 2014, the host venue for the finalEvent2026 FIFA World CupDateJuly 19, 2026 (2026-07-19)VenueMetLife Stadium, East Rutherford, New Jersey ← 2022 2030 → The 2026 FIFA World Cup final will be the final match of the 2026 World Cup, the 23rd edition of FIFA's competition for men's national football teams. The match is scheduled to be played at MetLife Stadium at the Meadowlands Sports Complex in East Rutherford, New Jersey, near New York City, on July 19, 2026. Background[edit] FIFA announced the date of the final on March 16, 2023. [1] The host of the final, MetLife Stadium, was announced by FIFA on February 4, 2024. [2] The announcement was originally anticipated for late 2023, but was delayed amid planning difficulties.

Evidence 5
Source Title: The 2026 World Cup final will take place at New Jersey's MetLife Stadium
Content: will host the opener of the 104-game tournament at Estadio Azteca on June 11. During the event, though, the 82,500-capacity stadium will be officially referred to as the "New York New Jersey Stadium" to comply with the FIFA's policy against non-sponsor corporate names. It's a World Cup of firsts. For the first time, the tournament will expand to include 48 teams, up from the 32 team-format held for the past seven tournaments. It will also be the first time the tournament is staged across three host nations. Beyond the New York-New Jersey complex, 15 other major cities were picked to host the World Cup matches. Sponsor Message MetLife Stadium, home to the New York Jets and Giants, hosted the Super Bowl in 2014 and the Copa América Centenario final in 2016. New York City Mayor Eric Adams and New Jersey Gov. Phil Murphy celebrated the announcement on social media. FIFA World Cup 2022
The U.S. cities hosting the 2026 World Cup are announced
"As a lifelong soccer fan, I am thrilled to announce that the FIFA World Cup 2026 Final will be hosted by New Jersey and New York City!" Gov. Murphy said in a tweet.
Reasoning: The claim states that the 2026 FIFA World Cup final will be held at MetLife Stadium in New Jersey, USA. The content from the evidence confirms this. Evidence 1, 2, 3, 4, and 5 directly mention that FIFA announced MetLife Stadium in East Rutherford, New Jersey, as the venue for the final on July 19, 2026. There is no contradictory evidence in the search results, and all sources consistently support the claim. The evidence is specific, authoritative (e.g., ESPN, FIFA announcements, Wikipedia), and unambiguous. No part of the claim is challenged or left unverified.
Decision: \#\#\#supported\#\#\#\\

Claim: Mount Katahdin is 6,288.2 feet (1,917.6 meters) tall.
Searched Results: Evidence 1
Source Title: 8 Tips for Hiking Mount Katahdin: Learn From Our Mistakes
Content: Skip to content

Hiking Mount Katahdin is a goal for many outdoor enthusiasts! Whether they’re looking to complete the Appalachian Trail or hike the highest mountain in Maine, summiting Katahdin is their answer. Katahdin stands at 5,269 feet tall. Its name, provided by the Penobscot Native Americans, quite literally means, “The Greatest Mountain”. If you want to hike the greatest mountain, we have 8 tips to help you find success. Learn from our mistakes to make hiking Mount Katahdin your reality! Keep Maine Beautiful

Evidence 2
Source Title: 8 Tips for Hiking Mount Katahdin: Learn From Our Mistakes
Content: Skip to content Hiking Mount Katahdin is a goal for many outdoor enthusiasts! Whether they’re looking to complete the Appalachian Trail or hike the highest mountain in Maine, summiting Katahdin is their answer. Katahdin stands at 5,269 feet tall. Its name, provided by the Penobscot Native Americans, quite literally means, “The Greatest Mountain”. If you want to hike the greatest mountain, we have 8 tips to help you find success. Learn from our mistakes to make hiking Mount Katahdin your reality! Keep Maine Beautiful In a Hurry? Let us help! Plan Ahead \& Reserve Wake Up Early \& Wait Train for Your Hike Prepare for Exposure Knife Edge on Return High Energy Snacks Turn Around if Needed AT Hikers Galore 8 Tips for Hiking Mount Katahdin Here are 8 suggestions we have for anyone looking to hike Mount Katahdin. The Baxter beauty can be difficult to summit for a variety of reasons, learn from our mistakes and hike the Northern Terminus of the Appalachian Trail. Plan Ahead When Hiking Mount Katahdin to Reduce Stress! 1. Plan Ahead \& Reserve a Spot
The number of hikers who can climb Mount Katahdin is limited by the number of parking spots at the trailheads.

Evidence 3
Source Title: Katahdin - Peakbagger.com
Content: towering form, and to hikers in search of truly rugged mountain majesty. Mount Katahdin is special due to a variety of factors. It is not a simple mountain, but a broad massif of several peaks, cirques, and ridges, surrounded on almost three sides by a ring of lower summits. This concentrated group of mountains stands utterly alone in the otherwise flat Maine north woods, and the southern face of the main mountain mass rises directly 4,000 feet from the Penobscot River to the highest summit in the entire state. The remote, and, compared to other eastern mountains, almost primeval forest setting of the peak is also very alluring, as is the large area above timberline (about 3800 feet at 46 degrees north). And finally, the spectacular sawtoothed Knife Edge, a serrated crest dropping thousands of feet on both sides, gives Katahdin a special kind of alpine grandeur.

Evidence 4
Source Title: 8 Tips for Hiking Mount Katahdin: Learn From Our Mistakes
Content: event that you cannot, please don’t find ay shame in turning back. 8. Experiencing AT hikers! Hiking Mount Katahdin is the final push for North Bound Appalachian Trail Thruhikers. It’s likely you’ll experience them along the trail and on the summit. They’re incredibly neat individuals who have so many stories to tell; if they’re looking to share, you should take a listen! It’s also important to realize that they’re probably going to smell… like really bad. And that’s okay! You would smell too after hiking 2,190 miles over the course of 5-7 months. 10 Frequently Asked Questions About Hiking Mount Katahdin 1. What is the Best Route Up Katahdin? There are several options for hiking Mount Katahdin. The Easiest, 10.4 Mi. : The Saddle Trail may be considered the easy route, however, it’s still no joke. You still need to gain thousands of feet in elevation, however, you’re able to do so in more mileage. The Shortest, 7.3 Mi. : The Abol Trail is the most direct trail to the summit of Mount Katahdin. It’s also the steepest route, gaining 3,982 feet of elevation in 3.4 miles. Keep in mind that the shortest trail isn’t necessarily the easiest.

Evidence 5
Source Title: Katahdin - Peakbagger.com
Content: Peaks (Rank \#1) Maine 3,500-foot Peaks (Rank \#1) Maine Peaks with 1000 feet of Prominence (Rank \#1) (Peak is on over 20 lists; Not all shown here.) Nearby Peak Searches: Radius Search - Nearest Peaks to Katahdin Elevation Ladder from Katahdin Prominence Ladder from Katahdin Description: Katahdin! The name of perhaps the single most outstanding peak in all the Appalachians is a magic word to Appalachian Trail through hikers who walk for 2,000 miles to reach it, to rock-climbers who challenge its rocky walls, to tourists who gape at its towering form, and to hikers in search of truly rugged mountain majesty.

Reasoning: The claim states that Mount Katahdin is 6,288.2 feet tall. However, Evidence 1 and 2 explicitly state that Katahdin "stands at 5,269 feet tall", refuteing the claim. Evidence 3 and 5 describe the mountain’s topography but do not mention its height, and Evidence 4 discusses elevation gain on trails but does not provide the summit’s total height. None of the evidence supports the claimed height of 6,288.2 feet. Since the explicit height provided in the search results (5,269 feet) refutes the claim and no other evidence supports the claimed height, the claim is refuted.
Decision: \#\#\#refuted\#\#\#\\

Claim: Taral Hicks released a cookbook in 2012.
Searched Results: Evidence 1
Source Title: Taral Hicks
Content: 100 Centre Street - Episode: "Fathers" 2003 Soul Food Naomi Episode: "The New Math" 2018 Illusions Kelly Main Cast 2020 Chase Street Beverly Johnson Main Cast References[edit] \^{} Jump up to: a b Beckerman, Jim (August 19, 2000). "Where Stars Are Born". The Record (North Jersey). Archived from the original on May 16, 2011. Retrieved February 12, 2020. When Shanell Jones graduated from Teaneck High School in June, she already had a deal with Def Jam, a major recording label. But as former Motown Records artist Taral Hicks (Teaneck, Class of 1994) and Alligator recording artist Shemekia Copeland (Teaneck, Class of 1997) could tell her, that's no big deal in this neck of the woods. \^{} "Tyler Perry's Aunt Bam's Place Starring Paris Benet \& Taral Hicks Comes to DVD Tomorrow". Urbanbridgez.com. June 11, 2012. Retrieved February 15, 2019. \^{} "Taral Hicks: From the Belly of the Beast". All Hip Hop.com. September 1, 2009. Archived from the original on September 4, 2009. \^{} Hicks, Taral. "Happy 20th V-Day my love @lorendawson". Instagram.com. Retrieved February 23, 2019. \^{} "Billboard – Music Charts, News, Photos \& Video". Retrieved May 15, 2017.

Evidence 2
Source Title: Taral Hicks
Content: Main Cast References[edit] \^{} Jump up to: a b Beckerman, Jim (August 19, 2000). "Where Stars Are Born". The Record (North Jersey). Archived from the original on May 16, 2011. Retrieved February 12, 2020.

Taral Hicks at IMDb
Taral Hicks at AllMusic

Categories:

1974 births
Actresses from the Bronx
20th-century African-American women singers
20th-century American women singers
20th-century American singers
American film actresses
African-American women singer-songwriters
American women singer-songwriters
American contemporary R\&B singers
Living people
Teaneck High School alumni
Singer-songwriters from New Jersey
Singers from New York City
African-American actresses
American television actresses
Singer-songwriters from New York (state)
Actresses from Teaneck, New Jersey
Hidden categories:

Articles with short description
Short description is different from Wikidata
Use mdy dates from March 2021
Articles with hCards
Pages using infobox musical artist with associated acts
All articles with unsourced statements
Articles with unsourced statements from February 2021

Evidence 3
Source Title: Taral Hicks
Content: Jeffrey Lewis, Maurice Lauchner, and Melonie Daniels. It had a 3-day run beginning August 30, 2011 in Atlanta, Georgia (Cobb Energy Center) and was filmed for a DVD release on June 12, 2012. [2] Music[edit] In 1995, Hicks signed a deal with Motown Records and released an album titled This Time. The single "Ooh, Ooh Baby", written by and featuring Missy Elliott, charted on the Billboard R\&B singles chart. However, the lead single intended to debut her singing career was "Distant Lover", an uptempo track produced by Teddy Riley.
Personal life[edit]
Hicks is a 1993 graduate of Grace Dodge Vocational High School in the Bronx, New York.

Evidence 4
Source Title: | Eat Your Books
Content: Popular Books Magazines Blogs Online Recipes Connect Gift Memberships Forum EYB Blog Email Preferences Press Get our Newsletter Facebook Bluesky Pinterest About |Contact |Terms of Use |Privacy |Report an Error X Already a Member? Sign In Email or Username Password Keep me Signed In New here? Become a Member And you have a search engine for ALL your recipes! Your cookbooks become searchable Your magazines become searchable Save online recipes in one place Chat with other cookbook lovers And you have a search engine for ALL your recipes!

And you have a search engine for
ALL your recipes!

Reasoning: The claim states that Taral Hicks released a cookbook in 2012. The evidence provided (Evidence 1-3) focuses on her acting career, music, and personal background (e.g., roles in TV/film, her 1995 album, education). Evidence 4 is unrelated to Taral Hicks and discusses a cookbook indexing service. None of the sources mention a cookbook authored by Taral Hicks in 2012 or any year. The claim’s core entity ("Taral Hicks") is clear, but the specific assertion about a cookbook lacks any supporting evidence. There is also no contradictory evidence explicitly stating she did not release a cookbook. The absence of relevant information in the evidence means there is not enough evidence to support or refute the claim.
Decision: \#\#\#not enough evidence\#\#\#\\

Claim: Alma Katsu spent her childhood there.
Searched Results: Evidence 1
Source Title: Alma Katsu - Book Series In Order
Content: she was a technology expert. Since 2012, she has been a consultant with the RAND Corporation where she holds the title of senior policy analyst. Katsu currently lives in a suburb of Washington with Bruce Katsu her musician husband. In addition to writing fiction, she also contributes to the Huffington Post and writes reviews for Publishers Weekly. Katsu was born to a Japanese born mother and an American born father who lived in Fairbanks Alaska. Nonetheless, she spent much of her childhood and young adult years in Concord Massachusetts, and attributes her life there as one of the reasons why she has such an interest in early American history. She graduated from Brandeis University in 1981 with a Bachelors in Writing and Literature from a class that included Margaret Rey the children’s book author and novelist John Irving. In 2004, she attended Johns Hopkins University from which she graduated with a Masters in Fiction. Outside the formal education setting, she attended the Squaw Valley writing workshops, which gave her the courage to finally become a published author. Alma Katsu’s works are generally known for the quality prose and depiction of supernatural settings in a realistic and immediate way.

Evidence 2
Source Title: Alma Katsu - Book Series In Order
Content: policy analyst. Katsu currently lives in a suburb of Washington with Bruce Katsu her musician husband. In addition to writing fiction, she also contributes to the Huffington Post and writes reviews for Publishers Weekly. Katsu was born to a Japanese born mother and an American born father who lived in Fairbanks Alaska. Nonetheless, she spent much of her childhood and young adult years in Concord Massachusetts, and attributes her life there as one of the reasons why she has such an interest in early American history. She graduated from Brandeis University in 1981 with a Bachelors in Writing and Literature from a class that included Margaret Rey the children’s book author and novelist John Irving. In 2004, she attended Johns Hopkins University from which she graduated with a Masters in Fiction. Outside the formal education setting, she attended the Squaw Valley writing workshops, which gave her the courage to finally become a published author. Alma Katsu’s works are generally known for the quality prose and depiction of supernatural settings in a realistic and immediate way. Her debut novel “The Taker” is a novel with a setting in the past which nevertheless has a modern day narrative as a frame of reference.

Evidence 3
Source Title: Alma Katsu
Content: 22, 2025. External links[edit] Wikimedia Commons has media related to Alma Katsu. Official Website: Facebook page: Alma Katsu at the Internet Speculative Fiction Database "Katsu Awards". Science Fiction Awards Database. Mark R. Kelly and the Locus Science Fiction Foundation. Simon \& Schuster website Italian publisher's website: https://web.archive.org/web/20130518005614/http://www.longanesi.it/scheda-autore.asp?editore=Longanesi\&idautore=4273 Brazilian publisher's website: http://www.editoranovoconceito.com.br/autores/alma-katsu hide Authority control databasesCategories: Living people 1959 births American women novelists American novelists of Asian descent American women novelists of Asian descent American women writers of Asian descent 21st-century American novelists 21st-century American women writers Novelists from Alaska Writers from Fairbanks, Alaska American writers of Japanese descent Brandeis University alumni Johns Hopkins University alumni RAND Corporation people Hidden categories: Webarchive template wayback links Articles with short description Short description is different from Wikidata Use mdy dates from December 2022 Commons category link from Wikidata This page was last edited on 1 April 2025, at 00:47 (UTC). Text is available under the Creative Commons Attribution-ShareAlike 4.0 License; additional terms may apply. By using this site, you agree to the Terms of Use and Privacy Policy.
Reasoning: The claim refers ambiguously to “there” without specifying the location within the claim itself. Although the evidence mentions Concord, Massachusetts, one cannot use the evidence to disambiguate an undefined referent in the claim. Because the claim’s core element (“there”) lacks a clear, self-contained referent, it is impossible to meaningfully verify or refute the statement.
Decision: \#\#\#unverifiable\#\#\#\\

Your task:\\

Claim: Kami encompass natural elements, ancestors, and abstract forces.
Searched Results: \{Retrieved Evidence from Scraped Webpages\}
Reasoning: 
\end{tcolorbox}

\section{Prompts for Failure Type Analysis}
\label{appdx:prompt-error}
The prompts for analyzing the distribution of claim extraction failures in Appendix \ref{appdx:failure-distribution-analysis-of-veriscore} are displayed below. This includes the prompt for detecting failures in existing extracted claims (\ref{prompt:unverifiable-redundant-claims}), and the prompt for detecting failures in possible missing claims (\ref{prompt:missing-claims}).
\subsection{Detection of Unverifiable and Redundant Claims}
\label{prompt:unverifiable-redundant-claims}

\begin{tcolorbox}[breakable]
\obeylines

\small
Fact-checking involves assessing the veracity of a given text, which could be a statement, claim, paragraph, or a generated response from a large language model.
For more complex fact-checking tasks, the input text is often broken down into a set of manageable, verifiable and self-contained sub-claims, a process called 'Input Decomposition'. 
Each sub-claim is required to be *self-contained*, meaning it is completely interpretable without the original text and other sub-claims. 
Each sub-claim is verified independently, and the results are combined to assess the overall veracity of the original input.
The decomposition quality is evaluated by checking whether each sub-claim meets this standard.
If not, it is marked Problematic and assigned an error type.\\

The decomposition error categories are defined as below:
\# Error Categories in Decomposition
—
\#\#\# Unverifiable Claims
Unverifiable claims are statements that do not describe a single event or state with all necessary modifiers (e.g., spatial, temporal, or relative clauses).
This category encompasses three sub-categories:\\

\#\#\#\# Subjective
- **Definition**: Claims about a story, personal ex-periences, hypotheticals (e.g., "would be" or subjunctive), subjective statements (e.g., opinions), future predictions, suggestions, advice, instructions, etc.\\

\#\#\#\# Tautology
- **Definition**: Claims that carry no meaning.\\

\#\#\#\# Ambiguous
- **Definition**: Claims that contains equivocal information or that are contextually dependent (e.g., not situated within relevant temporal information and location, or no clear referent for entities).\\

\#\#\# Redundant Claims
Redundant claims are repeated claims that cover identical information.
This category encompasses two sub-categories:\\

\#\#\#\# Intra-sentence redundancy
- **Definition**: overlapping claims extracted from the same sentence.\\

\#\#\#\# Inter-sentence redundancy
- **Definition**: Overlapping claims extracted across different sentences.\\

\#\#\# Task
Your task is to evaluate each sub-claim individually as either 'Acceptable' or 'Problematic'.
Please give your final judgment and support it with your justification. 
If a sub-claim is problematic, identify the error category it belongs to.
You must first report the main category, followed by the sub-category if applicable.\\

Use tripple backticks to enclose the reasoning process, and judgment.
You MUST return a separate Reasoning and Judgment block for each sub-claim.  
Each sub-claim should be presented in the following format:\\

Sub-claim i
\#\#\#\ Reasoning
'''
Your reasoning here.
'''\\

\#\#\# Judgment
'''
Conclude whether each sub-claim is 'Acceptable' or 'Problematic'.
If the sub-claim is problematic, identify the error category it belongs to.
The number of judgments MUST match the number of sub-claims in the Decomposition section, and be listed in the same order.
'''\\

Here are some examples:
\#\#\# Given text
'''
On the other hand, a posteriori knowledge relies on empirical evidence for justification.
'''\\

\#\#\# Decomposition
'''
1. A posteriori knowledge relies on empirical evidence.
2. A posteriori knowledge relies on empirical evidence for justification.
'''\\

Sub-claim 1  
\#\#\# Reasoning  
'''  
This sub-claim states that "a posteriori knowledge relies on empirical evidence."  
It is a well-formed, self-contained, and verifiable claim that accurately reflects the meaning of the original sentence.  
Although it omits "for justification," the core proposition about the nature of a posteriori knowledge remains clear and factual.  
It does not suffer from redundancy on its own and is interpretable without additional context.  
'''  \\

\#\#\# Judgment  
'''  
Acceptable  
'''  \\

Sub-claim 2  
\#\#\# Reasoning  
'''  
This sub-claim repeats the same core content as Sub-claim 1 but adds the phrase "for justification."  
The addition does not introduce a distinct factual proposition, and both claims are derived from the same sentence.  
Therefore, the two sub-claims overlap semantically and represent a form of intra-sentence redundancy.  
Including both introduces unnecessary duplication.  
'''  \\

\#\#\# Judgment  
'''  
Problematic: Redundant Claims - Intra-sentence redundancy  
'''  \\

==========

\#\#\# Given text
'''
For example, mathematical truths like "2+2=4" are known a priori and are considered to be necessarily true.
'''\\

\#\#\# Decomposition
'''
1. "2+2=4" is considered to be necessarily true.
2. A truth are considered to be necessarily true.
'''\\

Sub-claim 1
\#\#\# Reasoning
'''
This sub-claim is a specific and self-contained statement: `"2+2=4" is considered to be necessarily true.`  
It refers to a concrete mathematical proposition, and the claim about its modal status ("necessarily true") is a common and meaningful assertion in epistemology.  
The statement is syntactically and semantically correct and is verifiable within the context of analytic truths.  
There is no redundancy, ambiguity, tautology, or subjectivity involved.
'''\\

\#\#\# Judgment
'''
Acceptable
'''\\

Sub-claim 2
\#\#\# Reasoning
'''
The sub-claim "A truth are considered to be necessarily true." is both grammatically incorrect ("a truth are") and semantically empty.  
It repeats the notion that something already identified as "a truth" is "necessarily true" without adding any content.  
Because "truth" already entails truthfulness, asserting its necessity in this vague form is circular and does not convey any new verifiable information.  
This makes the claim a tautology — it carries no distinct or informative meaning and cannot be independently verified.
'''\\

\#\#\# Judgment
'''
Problematic: Unverifiable Claims - Tautology
'''\\

==========

Now, it is your turn to evaluate the following input decomposition.
\#\#\# Given text
'''
\{input\}
'''\\

\#\#\# Decomposition
'''
\{decomposition\}
'''
\end{tcolorbox}

\subsection{Detection of Missing Claims}
\label{prompt:missing-claims}

\begin{tcolorbox}[breakable]
\obeylines

\small
Fact-checking involves accessing the veracity of a given text, which could be a statement, claim, paragraph, or a generated response from a large language model.
For more complex fact-checking tasks, the input text is often broken down into a set of manageable, verifiable and self-contained sub-claim, a process called 'Input Decomposition'. 
Each sub-claim is required to be *self-contained*, meaning it is completely interpretable without the original text and other sub-claims. 
Each sub-claim is verified independently, and the results are combined to assess the overall veracity of the original input.
A high-quality decomposition must include **all verifiable claims** found in the original input. However, some valid claims may be missing from the decomposition.  
Such omissions are referred to as *Missing Claims* as defined below.\\

\#\#\# Error Category: Missing Claims
Missing claims are verifiable statements that appear in the input text but are not captured in the current decomposition.  
These errors may arise due to the following reasons:\\

\#\#\#\# Removed by relevance checks
- **Definition**: A valid sentence-level claim was removed post-hoc due to mistaken relevance filtering or heuristic revisions. \\

\#\#\#\# Missed by the extractor
- **Definition**: A valid claim is expressed across multiple sentences but was not extracted due to sentence-level processing limitations.\\

\#\#\# Task
Your task is to review the original input and the decomposition. Then, identify any *Missing Claims* that meet the definition above.  
If a claim is missing, specify it exactly and indicate which of the two error types it corresponds to.
You must first report the main category, followed by the sub-category.\\

Use tripple backticks to enclose the reasoning process, and judgment.
You MUST return a separate Reasoning and Judgment block for each missing sub-claim.\\  
Each missing sub-claim should be presented in the following format:\\

Missing Sub-claim i
\#\#\# Reasoning
'''
Your reasoning here.
Explain why this claim is considered verifiable and self-contained, why it is missing from the current decomposition, and which category of missing claim it belongs to (Removed by relevance check or Unextracted inter-sentence claim).  
'''  \\

\#\#\# Content of Missing Claim
'''
The content of the missing claim should be presented here.
'''\\

\#\#\# Judgment
'''
Missing Claim: <main category> - <sub-category> 
'''\\

Here are some examples:
\#\#\# Given text
'''
A priori knowledge refers to knowledge that is independent of experience, while a posteriori knowledge is knowledge that is derived from experience.
'''\\

\#\#\# Decomposition
'''
1. A priori knowledge refers to knowledge that is independent of experience.
'''\\

Missing Sub-claim 1
\#\#\# Reasoning  
'''  
The original sentence defines both a priori and a posteriori knowledge.
While the decomposition includes only the a priori part, it omits the definition of a posteriori knowledge, which is a parallel and equally verifiable factual statement. 
The a posteriori clause is syntactically complete, semantically distinct, and interpretable on its own. 
Since it was not included despite being a valid, sentence-contained claim, its omission likely results from a relevance filtering issue or early truncation.
'''\\

\#\#\# Content of Missing Claim
'''
A posteriori knowledge is knowledge that is derived from experience.
'''\\

\#\#\# Judgment  
'''  
Missing Claims - Removed by relevance checks
'''  \\

==========\\

\#\#\# Given text
'''
E-commerce Product Moderation:
Step 1: Automated Content Screening: [...more sentences...]
Step 2: Contextual Human Review: [...more sentences...]
'''\\

\#\#\# Decomposition
'''
1. E-commerce Product Moderation contains two steps of automated content screening and contextualized human review.
'''\\

Missing Sub-claim 1
\#\#\# Reasoning
'''
The input outlines Step 1: Automated Content Screening, and provides additional information (represented by "[...more sentences...]") describing what this step involves. These sentences likely contain concrete procedures or criteria, such as filtering specific categories of content, keyword matching, or automated rejection rules. Each of these is a verifiable, self-contained claim and should be extracted. Their absence reflects the extractor’s inability to handle multi-sentence procedural content.
'''\\

\#\#\# Content of Missing Claim
'''
Automated content screening involves predefined rules or filters to detect inappropriate product content.
'''\\

\#\#\# Judgment
'''
Missing Claims - Missed by the extractor
'''\\

Missing Sub-claim 2
\#\#\# Reasoning
'''
Step 2: Contextual Human Review is also elaborated through additional sentences (not shown in full). These likely describe review processes such as human evaluation of flagged items, applying contextual judgment, or reviewing based on cultural norms. Each of these processes is independently verifiable and interpretable. These were not included in the decomposition despite being factual and relevant. This qualifies as missed inter-sentence claims.
'''\\

\#\#\# Content of Missing Claim
'''
Contextual human review includes manual examination of flagged products using context-aware criteria.
'''\\

\#\#\# Judgment
'''
Missing Claims - Missed by the extractor
'''\\

==========\\

Now, it is your turn to review the given text and decomposition, and identify any missing claims.
\#\#\# Given text
'''
{input}
'''\\

\#\#\# Decomposition
'''
\{decomposition\}
'''
\end{tcolorbox}

\section{Ethics Statement}
\label{sec:ethics}
We recognize the importance of ethical considerations in our work. All instances included in the constructed benchmark are sourced from publicly accessible existing datasets, and no proprietary or confidential data were used. The associated claims were annotated through a rigorous process, with additional quality control by domain experts. Annotators were fully informed about the research purpose and provided their consent voluntarily. No personally identifiable or sensitive information was collected.  To mitigate potential misuse, particularly in downstream applications that might involve automated verification in policy or legal contexts, we emphasize that the benchmark is intended solely for academic research. Outputs from models trained or evaluated on the benchmark should not be used in isolation for critical decision-making. To support reproducibility and responsible use, we release the dataset, code, annotation tool, and documentation under a permissive research license upon publication. 

\end{document}